\documentclass[sigconf, nonacm]{acmart}

\AtBeginDocument{%
	\providecommand\BibTeX{{%
			\normalfont B\kern-0.5em{\scshape i\kern-0.25em b}\kern-0.8em\TeX}}}

\usepackage{graphicx}
\usepackage{subfigure}
\usepackage{hyperref}

\usepackage{amsmath}
\usepackage{amssymb}
\usepackage[export]{adjustbox}
\usepackage{arydshln} 
\usepackage{wrapfig}

\usepackage{hyperref}       
\usepackage{booktabs}       
\usepackage{amsfonts}       

\usepackage{natbib}
\usepackage{cuted}
\usepackage{placeins}
\usepackage{xcolor}

\newcommand{\bs}[1]{\boldsymbol{#1}}
\newcommand{\CD}{\mathcal{D}}
\newcommand{\EE}{\mathbb{E}}
\newcommand{\zv}{\boldsymbol{z}}
\newcommand{\thetav}{\boldsymbol{\theta}}
\newcommand{\xv}{\boldsymbol{x}}

\newcommand{\epsilonv}{\boldsymbol{\epsilon}}

\setcopyright{none}
\renewcommand\footnotetextcopyrightpermission[1]{} 
\setcopyright{none}
\makeatletter
\renewcommand\@formatdoi[1]{\ignorespaces}
\makeatother

\acmConference[CHIL'20]{The ACM Conference on Health, Inference, and Learning}{April 2020}
{Toronto, Canada}

\usepackage{balance}

\begin{document}
	
	\title{Survival Cluster Analysis}

	\author{Paidamoyo Chapfuwa}
	\affiliation{\institution{Duke University}}
	\email{paidamoyo.chapfuwa@duke.edu}
	
	\author{Chunyuan Li}
	\affiliation{\institution{Microsoft Research, Redmond}}
	\email{chunyl@microsoft.com}
	
	\author{Nikhil Mehta}
	\affiliation{\institution{Duke University}}
	\email{nm208@duke.edu}
	
	\author{Lawrence Carin}
	\affiliation{\institution{Duke University}}
	\email{lcarin@duke.edu}
	
	\author{Ricardo Henao}
	\affiliation{\institution{Duke University}}
	\email{ricardo.henao@duke.edu}

	\begin{abstract}
		Conventional survival analysis approaches estimate risk scores or individualized time-to-event distributions conditioned on covariates.
		In practice, there is often great population-level phenotypic heterogeneity, resulting from (unknown) subpopulations with diverse risk profiles or survival distributions. 
		As a result, there is an unmet need in survival analysis for identifying subpopulations with distinct risk profiles, while jointly accounting for accurate individualized time-to-event predictions. 
		An approach that addresses this need is likely to improve characterization of individual outcomes by leveraging regularities in subpopulations, thus accounting for population-level heterogeneity. 
		In this paper, we propose a Bayesian nonparametrics approach that represents observations (subjects) in a clustered latent space, and encourages accurate time-to-event predictions and clusters (subpopulations) with distinct risk profiles.
		Experiments on real-world datasets show consistent improvements in predictive performance and interpretability relative to existing state-of-the-art survival analysis models.
	\end{abstract}
	
	\maketitle

	\section{INTRODUCTION}
	Time-to-event models have primarily focused on either estimating a (point estimate) risk score or individualized time-to-event distributions.
	Parametric models estimate the time-to-event distribution conditional on covariates by assuming a parametric form of the event distribution, {\em i.e}, exponential, Weibull, log-normal, {\em etc}.
	Parametric models fall under the Accelerated Failure Time (AFT) \citep{SIM:SIM4780111409} framework, provided they assume covariates either accelerate or decelerate the time-to-event.
	Assuming a parametric distribution is inflexible as the hazard function depends on the selected baseline distribution, for example assuming an exponential distribution, yields a constant hazard rate function.
	Alternatively, Cox Proportional Hazards (CoxPH) \citep{cox1992regression}, a semi-parametric, linear model for estimating relative risks is widely used in practice, as it does not require one to specify the baseline distribution.
	For pre-specified time-horizons, the non-parametric Random Survival Forest (RSF) \citep{ishwaran2008random} was proposed to estimate the cumulative hazard function based on an ensemble of binary decision trees, albeit often limited by scaling problems for large and high-dimensional datasets.
	
	With recent advances in machine learning, deep learning methods have improved classical survival analysis methods by leveraging non-linear relationship between covariates, for improved time-to-event or risk score predictions.
	Deep learning methods inspired by CoxPH or AFT have been proposed, {\em e.g.}, DeepSurv \citep{katzman2018deepsurv}, Deep Survival Analysis (DSA) \citep{ranganath2016deep}, Deep Regularized Accelerated Failure Time (DRAFT) \citep{chapfuwa2018adversarial}, Gaussian-process-based models \citep{fernandez2016gaussian, alaa2017gaussianmulti}, and the Survival Continuous Ranked Probability Score (S-CRPS) \citep{avati2018countdown}.
	Sampling-based nonparametric methods have been proposed as well, {\em e.g.}, normalizing-flow-based DSA \citep{miscouridou2018deep}, adversarial-learning-based Deep Adversarial Time
	to Event (DATE) \citep{chapfuwa2018adversarial} and Survival Function Matching (SFM) \citep{chapfuwa2019sfm}.
	Another class of nonparametric methods discretize time-to-event to predict survival probability within pre-specified discrete-interval event times with logistic-regression-based methods \citep{yu2011learning, fotso2018deep, lee2018deephit}.
	Further, deep learning methods have also successfully addressed calibration \citep{chapfuwa2019sfm, avati2018countdown, lee2019temporal} and competing risks \citep{zhang2018nonparametric, alaa2017gaussianmulti}.
	
	\begin{figure*}%
		\centering
		\begin{tabular}{ccc}
			\includegraphics[width=0.315\linewidth]{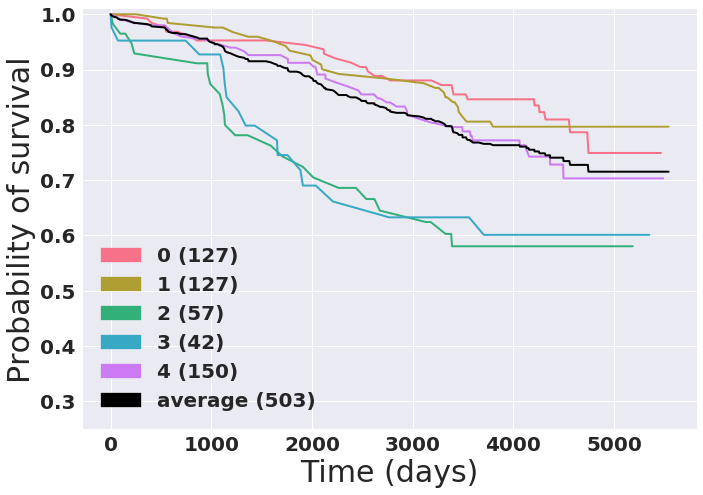}
			&
			\includegraphics[width=0.315\linewidth]{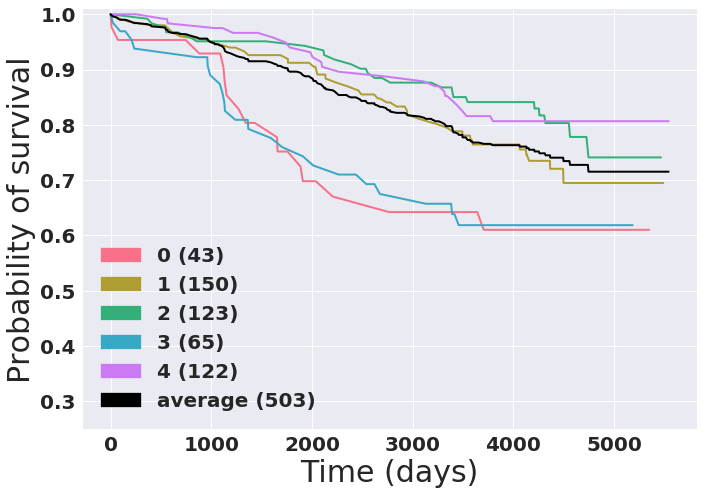}
			&
			\includegraphics[width=0.315\linewidth]{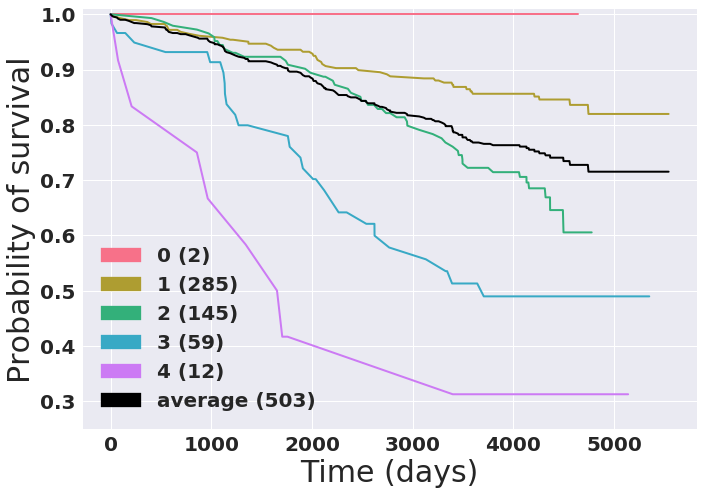}
			\\
			(a) $K$-means &
			(b) SSC-Bair &
			(c) SCA 
		\end{tabular}
		\vspace{-3mm}
		\caption{Cluster-specific Kaplan-Meier survival profiles for three clustering methods on the {\sc sleep} dataset (See Section~\ref{sc:exp} for details).
			(a) Standard $K$-means. (b) CoxPH-based covariate selection followed by $K$-means. (c) Proposed approach for joint learning of individualized time-to-event predictions and clustering.  By jointly learning clustering with respect to both the covariates $\xv$ and predicted time-to-event $t$, our model (SCA) can identify high-, medium- and low-risk individuals. Demonstrating  the need to account for time information via a non-linear transformation of covariates when clustering survival datasets. 
		}
		\label{fig:clustering}
	\end{figure*}
	
	Clustering based on risk-profiles in survival analysis is relatively under-explored in machine learning, but is critical in applications such as (clinical) decision making.
	Identifying phenotypically heterogeneous subpopulations in the context of risk prediction is an important step toward machine-learning-based models for precision medicine \citep{collins2015new, djuric2017precision}.
	Existing clustering methods for stratifying risks in survival analysis include feature based $K$-means (see Figure~\ref{fig:clustering}(a)) or hierarchical clustering \citep{eisen1998cluster,shah2015phenomapping,ahlqvist2018novel}.
	Principal component cluster analysis has also been considered \citep{ahmad2014clinical}.
	However, it is well understood that feature-based clustering in covariate space may produce clusters that are not consistent with survival outcomes \citep{bair2004semi, gaynor2013identification}, particularly for high-dimensional datasets, such as gene expression data.
	
	Methods that account for survival outcomes in clustering include CoxPH-inspired techniques \citep{bair2004semi, gaynor2013identification}, implemented as a two-step process: first, high CoxPH scoring covariates are selected, then a classical clustering approach like $K$-means is applied (see Figure~\ref{fig:clustering}(b)).
	However, CoxPH-based approaches are limited by the proportional hazards assumption.
	Alternatively, \citet{xia2019outcome} proposed an outcome driven attention-based multi-task deep learning model for classification and then applied $K$-means on the latent representations to cluster subjects with acute coronary syndrome.
	More recently, \citet{mouli2019deep} introduced DeepCLife, a method that learns clusters by maximizing the pairwise differences between the survival functions of all cluster pairs.
	This is done by indirectly maximizing the logrank score \citep{mantel1966evaluation}.
	Unlike DeepCLife, which aims to optimize clusters but not predictions, our goal is to jointly characterize time-to-event predictive distributions from a clustered latent space conditioned on covariates (see Figure~\ref{fig:clustering}(c)).
	
	We propose a model for time-to-event predictions equipped with a structured latent representation that allows for clustering via a prior for infinite mixture of distributions.
	We circumvent the challenges associated with infinite mixtures in stochastic learning by leveraging a truncated Dirichlet process (DP) with a stick breaking representation.
	The proposed model, termed Survival Clustering Analysis (SCA), is specified as: $i$) a deterministic encoder that maps covariates into a latent representation; $ii$) a stochastic time-to-event predictor that feeds from the latent representation; and $iii$) a distribution matching objective that encourages latent representations to behave as a mixture of distributions following a DP structure. 
	This approach allows identification and analysis of phenotypically heterogeneous subpopulations.
	Our experiments demonstrate that SCA yields consistent improvements in predictive performance and cluster quality relative to existing methods.
	
	\vspace{-2mm}
	\section{BACKGROUND}
	In a conventional time-to-event (survival analysis) setup, we are given $N$ observations.
	Individual observation are described by triplets $ \CD = \{ (\xv_i, t_i, l_i)\}_{i=1} ^N$, where $\xv_i = \{x_i, ..., x_p\}$ is a  $p$-dimensional vector of covariates, $t_i$ is the time-to-event and $l_i \in \{0, 1\}$ is the censoring indicator.
	When $l_i =0$ (censored) the subject has \emph{not} experienced an event up to time $t_i$, while $l_i=1$ indicates observed (ground truth) event times.
	
	Time-to-event models are conditional on covariates: the event time density function $f(t|\bs{x})$, the hazard rate (risk score) function $h(t|\xv)$ or the survival function $S(t|\xv) =  \ P(T>t) =  1-F(t|\xv)$, also known as the probability of failure occurring after time $t$, where $F(t|\xv)$ is the cumulative density function.
	From standard survival function definitions \citep{kleinbaum2010survival}, the relationship between these three characterizations is formulated as $f(t|\xv)=h(t|\xv)S(t|\xv)$.
	
	In practice, modern (often large) datasets are not homogeneous but composed of phenotipically heterogeneous subpopulations, {\em i.e.}, subsets of observations that cluster according to both covariates and time-to-event similarities.
	In a clinical setting for instance, identification of, {\em e.g.}, high-, medium- and low-risk subpopulations that are equipped with accurate estimates of time-to-event has the potential to result in a more cost effective way of targeting interventions, treatments or care delivery.
	We formulate an approach to jointly learn individualized time-to-event distributions and clusters informed by time-to-event profiles.

	\vspace{-2mm}
	\section{SURVIVAL CLUSTER ANALYSIS}\label{sc:sca}
	
	\begin{figure*}[t!]
		\centering
		\includegraphics[width=0.85\linewidth]{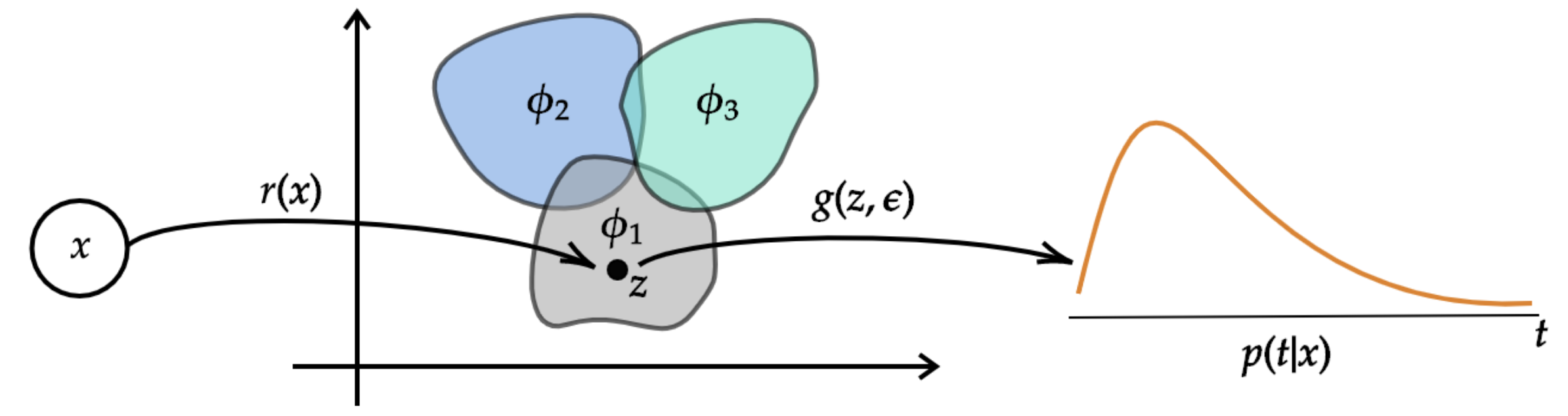}
		\caption{Illustration of Survival Clustering Analysis (SCA). The latent space has a mixture-of-distributions structure, illustrated as three mixture components $\{\phi_k\}_{k=1}^3$. Observation $\bs{x}$ is mapped into its latent representation via a deterministic encoding $\bs{z} = r_{\bs\psi}(\xv)$ belonging to $\phi_1$, which is then used to stochastically predict (via sampling) the time-to-event via $t = g_{\bs\theta}(\zv,\epsilonv)$.}
		\label{fg:model}
	\end{figure*}
	
	The Bayesian nonparametrics approach formulated below encourages latent representations to behave as a mixture of distributions, following a Dirichlet Process (DP) structure via a distribution matching approach. Further, we learn to cluster the latent space in a stochastic manner for which the number of clusters is unknown.
	To demonstrate the efficacy of our clustering algorithm, we also present a time-to-event prediction formulation, leveraging current state-of-the-art time-to-event prediction models.
	See the Supplementary Material for the list of variable definitions used in our formulation.
	
	\subsection{Clustering with Dirichlet Process}
	A DP is formally defined as $G \sim {\rm DP}(\gamma_{o}, G_o)$ and parametrized by the \emph{base probability measure} $G_o$ and \emph{concentration parameter} $\gamma_o >0 $~\citep{ferguson1973bayesian}.
	With probability one \citep{sethuraman1994constructive}:
	\begin{align}\label{eq:DP}
		G = \sum_{k=1}^{\infty} \pi_k \delta_{\phi_k}, \qquad \pi_k  = V_k \prod_{l=1}^{k-1} (1 - V_l) \,,
	\end{align}
	where $\phi_k \sim G_o$, $ \delta_{\phi_k}$ represents a probability measure concentrated at $\phi_k$ and $V_k \sim {\rm Beta}( 1, \gamma_o)$ are \emph{stick-breaking} weights with statistics that depend on parameter $\gamma_0$.
	The sequence $\bs{\pi} = \{\pi_k\} _{i=1} ^{\infty}$ satisfies $\sum_{k=1}^{\infty}  \pi_k =1$, with probability one, such that $\bs{\pi} \sim {\rm GEM}(\gamma_{o})$ \citep{pitman2002poisson}.
	Further, note that $\pi_k$ represents the likelihood that $G=\phi_k$.
	
	Discrete distribution $G$ is suitable as a prior for mixture components in infinite mixture models \citep{rasmussen2000infinite}.
	Further, the stick-breaking process \citep{sethuraman1994constructive} that generates $\bs{\pi}$ results in a mechanism that allows one to learn the number of mixture components (clusters) from data.
	In fact, the number of distinct atoms, $\{\phi_k\}_{k=1}^\infty$, has been shown to grow with the size of the data as $O(\log N)$ \citep{abramowitz1988handbook}.
	So motivated, we assume that data embedded in a latent space are distributed according to a mixture of distributions with parameters specified by the \emph{base probability measure} $G_o$, as described below.
	
	Assuming  exchangeable latent representations $\{\bs{z}\}_{i=1} ^{N}$, we propose generating event times following the generative process below
	\begin{align}
		p(\bs{c}) & = \sum_{k=1}^\infty \pi_k\delta_{\bs{c}_k} \label{eq:pc} \\ 
		\bs{z}_n & \sim st(\bs{c}_{u_n},\nu) \label{eq:pz} \\
		t_n & \sim g_{\bs\theta}(\bs{z}_n,\bs{\epsilon}_n)\,, \label{eq:pt} 
	\end{align}
	where $g_{\bs\theta}(\bs{z},\epsilon)$ is a function that implicitly represents the conditional time-to-event density, $f(t|\bs{x})$, specified as a neural network with parameters $\bs{\theta}$.
	The source of stochasticity, $\bs{\epsilon}$, for $g_{\thetav}(\bs{z},\bs{\epsilon})$, is set to a simple distribution $\bs{\epsilon}\sim p_{\bs{\epsilon}}$, {\em e.g},  uniform or Gaussian.
	The latent representation for the $n$-th observation, $\bs{z}_n$ is distributed according to $\phi_{u_n}=st(\bs{c}_{u_n},\nu)$, where $u_n$ is the mixture component membership indicator for $\bs{z}_n$.
	Lastly, together with \eqref{eq:pz}, $p(\bs{c})$ in \eqref{eq:pc} represents an infinite mixture of Student's $t$-distributions with $\nu$ degrees of freedom and means $\{\bs{c}_k\}_{k=1}^\infty$, each of which is drawn independently from the base probability measure $G_o$ as $\bs{c}_k\sim \mathcal{N}(\bs{0},\bs{I})$.
	
	The Student's $t$ distribution in \eqref{eq:pz} is a general yet parametrically simple distribution, robust to outliers and amenable to efficient computations and gradient estimates.
	It has been widely used in machine learning for mixture modeling \citep{rasmussen2000infinite}, clustering \citep{xie2016unsupervised} and visualization \citep{maaten2008visualizing}.
	Further, we formulate the $t$ distribution according to the normal-inverse-gamma likelihood, where  marginalizing out the variance yields a Student-$t$ distribution, see \citep{bishop2006pattern} for details.
	Interestingly, as special cases, when $\nu=1$, $\bs{z}_n$ is Cauchy distributed while for $\nu > 3$, $\bs{z}_n$ approaches a Gaussian distribution.
	
	The generative process above further requires learning a mapping function from covariates to latent space, $\bs{z}_n=r_{\bs\psi}(\bs{x}_n)$ with parameters $\bs{\psi}$, that is globally consistent with the mixture model prior in \eqref{eq:pc} and \eqref{eq:pz}, parameterized by $\{\pi_k,\bs{c}_k\}_{k=1}^\infty$.
	In addition, we also need to learn the parameters $\bs{\theta}$ of the time-to-event generating function $g_{\bs\theta}(\bs{z}_n,\epsilon_n)$ in \eqref{eq:pt}.
	This specification, illustrated in Figure \ref{fg:model}, constitutes the proposed Survival Clustering Analysis (SCA).
	
	Note that unlike existing unsupervised and supervised autoencoding approaches \citep{vincent2010stacked, kingma2013auto, jiang2016variational}, we do not seek to model the covariates, $\bs{x}$. Rather, we make time-to-event predictions based on a latent representation specified as a function of observed covariates, required to be consistent with a mixture of distributions prior.
	Consequently, we need not specify a decoding arm to reconstruct the covariates, $\bs{x}$.
	
	In practice, learning the mixture component assignments $u_n$ and a potentially infinite number of mixture components with minibatches (stochastically) is challenging, because the former constitutes a discrete random variable and the latter requires keeping track of the number of non-empty mixture components during learning.
	To circumvent this, we learn the mixture component assignments probabilistically as $q(u_n=k|\bs{x}_n)$, and use a truncated representation of the DP formulation \citep{ishwaran2001gibbs, blei2006variational}, which for large enough truncation number, denoted as $K$, is virtually indistinguishable from a standard DP \citep{ishwaran2001gibbs}.
	
	\subsection{Latent-Space Representation}
	Following the conventional maximum likelihood formulation for mixture models \citep{bishop2006pattern}, we can approximate the distributions for the mixture assignments and mixture proportions as follows
	\begin{align}
		q(u_n=k|\bs{x}_n) & = \frac{\alpha_{nk}}{\sum_{k=1}^{K}  \alpha_{nk}} \notag \\
		\alpha_{nk} & \propto \pi_k p(r_{\bs\psi}(\bs{x}_n)|\bs{c}_k,\nu) \notag \\
		q(\bs{\pi}|\bs{\xi},\{\bs{x}_1\}_{n=1}^M) & = {\rm Dir}(\bs{\pi}|\bs{\xi}) \label{eq:qpi} \\
		\xi_k & = \frac{1}{K} + \sum_{n=1}^M q(u_n=k|\bs{x}_n) \,, \notag
	\end{align}
	where $\bs{\xi}=\{\xi_k\}_{k=1}^K$ is a variational parameter for expectation $\mathbb{E}[\bs{\pi}]$, $M$ is the minibatch size and we have replaced $\bs{z}_n$ in \eqref{eq:pz} with the encoding of covariates into latent space, {\em i.e.}, $\bs{z}_n=r_{\bs\psi}(\bs{x}_n)$.
	However, \eqref{eq:qpi} is not necessarily consistent with the DP in \eqref{eq:pc} and its stick-breaking prior, $\bs{\pi}\sim{\rm GEM}(\gamma_0)$, which from \eqref{eq:DP} should result in
	\begin{align}
		p(u_n=k|\bs{x}_n) & = \frac{\beta_{nk}}{\sum_{k=1}^{K}  \beta_{nk}} \notag \\
		\beta_{nk} & \propto V_k \prod_{l=1} ^ {k-1} (1 - V_l) p(r_{\bs\psi}(\bs{x}_n)|\bs{c}_k,\nu) \notag \\
		p(\bs{\pi}|\bs{\gamma},\{\bs{x}_n\}_{n=1}^M) & = {\rm Dir}(\bs{\pi}|\bs{\gamma}) \label{eq:ppi} \\
		\gamma_k & = \gamma_0 + \sum_{n=1}^M p(u_n=k|\bs{x}_n) \,, \notag
	\end{align}
	where $V_k \sim {\rm Beta}(1,\gamma_{0})$, which in practice is complicated by the need to sample from the mixture proportion weights $\{V_k\}_{k=1}^K$.
	In our implementation, instead of sampling from $V_k$, we use its expectation, {\em i.e.}, $\mathbb{E}[V_k]=(1+\gamma_0)^{-1}$.
	Alternatively, we could also use a reparameterizable distribution such as the Kumaraswamy distribution, which is closely related to the Beta distribution as in \citep{nalisnick2016stick}. However, we found that using expectations, which is common in variational formulations \citep{blei2006variational, jordan1999introduction}, works well in practice.
	
	In order to make $q(\bs{\pi}|\bs{\xi},\{\bs{x}_1\}_{n=1}^M)$ in \eqref{eq:qpi} and $p(\bs{\pi}|\bs{\gamma},\{\bs{x}_n\}_{n=1}^M)$ in \eqref{eq:ppi} consistent, we want their distributions to match, {\em i.e.}, we seek to learn $\bs{\psi}$ of $\bs{z}_n=r_\psi(\bs{x}_n)$, $\{\pi_k\}_{k=1}^K$ and $\{\bs{c}_k\}_{k=1}^K$, so the approximation $q(\bs{\pi}|\bs{x}_1,\ldots,\bs{x}_N)$ matches the desired stick breaking behavior of \eqref{eq:ppi}.
	For this purpose, we minimize
	\begin{align}\label{eq:clust_obj}
		\begin{aligned}
			& \ell_{{\rm dp}}(\bs{\psi},\{\bs{c}_k\}_{k=1}^K; \CD) = \\
			& \hspace{16mm}{\rm KL} \left( q( \bs{\pi} | \bs{\xi},\{\bs{x}_n\}_{n=1}^M ) || p( \bs{\pi} | \bs{\gamma},\{\bs{x}_n\}_{n=1}^M) \right) \,.
		\end{aligned}
	\end{align}
	%
	The KL Divergence between the two Dirichlet distributions $q$ and $p$ with respect to their corresponding parameters $\xi$ and $\gamma$, has a desirable closed form formulation defined as
	\begin{align}
		\text{KL} \left(q||p) \right) & = \ln \Gamma (\xi_0) - \ln \Gamma (\gamma_0)  + \sum_{k=1} ^{K} \left( \ln \Gamma (\gamma_{k}) - \ln \Gamma (\xi_{k}) \right) \notag 
		\\
		& + \sum_{k=1}^{K} ( \xi_{k} - \gamma_{k})  \left( \Phi (\xi_{k}) -  \Phi(\xi_0) \right) \,,
	\end{align}
	where $\xi_0=\sum_{k=1}^K \xi_k$, $\gamma_0=\sum_{k=1}^K \gamma_k$, $\Phi(\cdot)$ is the \emph{digamma function} and $\Gamma(\cdot)$ is the \emph{Gamma function}.
	
	This loss function is used during learning to update $\bs{\psi}$ and $\{\bs{c}_k\}_{k=1}^K$. For $\{\pi_k\}_{k=1}^K$, the mixture proportions, we use a simple updating procedure akin to online expectation-maximization (EM) \citep{cappe2009line}.
	In particular, we update iteratively as
	\begin{align}\label{eq:c_update}
		\bs{\pi} ^{t+1} &= \eta  \bs{ \pi}^{t} + (1- \eta)\mathbb{E}[q( \bs{\pi} | \bs{\xi},\{\bs{x}_n\}_{n=1}^M )],
	\end{align}
	where $ 0 <\eta < 1$ is the step size and we initialize $\bs{\pi}_k ^0=1/K$.
	In practice, we set $\eta=0.9$; however, $\eta$ can also be selected using grid search.
	The online approach in \eqref{eq:c_update} is widely used to update global parameters in stochastic learning procedures.
	For instance, it has been used to learn the population mean and variance in batch normalization \citep{ioffe2015batch}.
	
	\subsection{Time-to-Event Distributions}
	In addition to the clustered, mixture representation of the latent space, we also seek a high-performing time-to-event model that yields concentrated, accurate and calibrated time-to-event predictions, while accounting for censored event times ($l_n = 0$).
	We borrow the accuracy objective from DATE \citep{chapfuwa2018adversarial} and the calibration objective from SFM \citep{chapfuwa2019sfm}.
	Below we describe these objectives in the context of our formulation.
	
	\paragraph{Accuracy Objective}
	The dataset $\mathcal{D}$ is split into two disjoint sets $(t, \xv) \sim p_c$ and $(t, \xv) \sim p_{nc}$, where $p_c$ and $p_{nc}$ represent censored and non-censored empirical distributions for these sets, respectively.
	We leverage the accuracy objective from DATE \citep{chapfuwa2018adversarial} formulated as
	\begin{align}\label{eq:date}
		\ell_{\rm acc}(\thetav,\bs{\psi};\mathcal{D}) & =\mathbb{E}_{(t,\xv)\sim p_{c},\epsilonv\sim p_\epsilon}[\max\left(0, t-g_{\thetav}(r_{\bs\psi}(\bs{x}), \epsilonv)\right)] \nonumber\\ 
		& +   \EE_{(t, \xv) \sim p_{nc},\epsilonv\sim p_\epsilon} [||t- g_{\thetav}\left(r_{\bs\psi}(\bs{x}), \epsilonv\right)||_1]\,,
	\end{align}
	where $\epsilonv\sim p_{\bs\epsilon}$ has a simple distribution (uniform or Gaussian). $\ell_{\rm acc}(\thetav;\mathcal{D})$ encourages that time-to-event samples from the model, evaluated on censored observations, $l_n=0$, are larger than the censoring time, while close to the ground truth for non-censored (observed) events, $l_n=1$.
	
	\begin{table*}
		\caption{Summary statistics of the datasets used in the experiments. The time range, $t_{\rm max}$, is noted in days except for {\sc seer} for which time is measured in months.}
		\label{tb:data}
		\vspace{-4mm}
		\centering
		\resizebox{0.65\linewidth}{!}{
			\begin{tabular}{lrrrrrr}
				& {\sc ehr} & {\sc flchain} & {\sc support} & {\sc seer}  & {\sc sleep} &  {\sc framingham}\\
				\toprule
				Events (\%) & 23.9 & 27.5 & 68.1 & 51.0  & 23.8 &  11.14\\
				$N$ & 394,823 & 7,894 & 9,105 & 68,082 & 5026  & 40,078\\
				$d$ (categorical) & 729 (106) & 26 (21) & 59 (31) & 789 (771) & 206&  12 (8) \\
				Missing (\%) & 1.9 & 2.1 & 12.6 & 23.4  & 18.2  & 0.33\\
				$t_{\rm max}$ & 365 & 5,215 & 2,029 & 120 & 5,794& 6,000  \\
				\bottomrule
			\end{tabular}
		}
	\end{table*}
	
	\paragraph{Calibration Objective}
	We desire that samples generated from the model $g_{\thetav}(r_{\bs\psi}(\bs{x}),\bs{\epsilon})$ match the empirical marginal distribution $p(t)$.
	We borrow the calibration objective from SFM \citep{chapfuwa2019sfm} defined over the set of distinct and ordered observed event times (censored and non-censored), $\mathcal{T} = \{ t_i |  t_i > t_{i-1} > \ldots > t_0\}$, 
	\begin{align}\label{eq:km_emd}
		\ell_{\rm cal}(\thetav,\bs{\psi};\CD) = \frac{1}{|\mathcal{T}|} \sum_{t_i \in \mathcal{T} } \left\| \hat{S}_{\rm PKM}^{p(t)}(t_i) - 	\hat{S}_{\rm PKM}^{g_{\thetav}(r_{\bs\psi}(\bs{x}),\bs{\epsilon})}(t_i) \right\|_1 \,,
	\end{align}
	where $\hat{S}_{\rm PKM}$ is formulated as:
	\begin{align}\label{eq:km_ref_est_step}
		\hspace{-1mm}\hat{S}_{\rm PKM}(t_i) & = \left(1- \frac{\sum_{n:l_n=1} H(\hat{T}_n - t_{i-1}) - H(\hat{T}_n  - t_i)}{ M  - \sum_{n=1}^{M} H(t_{i-1} -\hat{T}_n ) } \right) \notag \\ 
		& \times \hat{S}_{\rm PKM}(t_{i-1}) \,,
	\end{align}
	%
	%
	and $H(b) = \tfrac{1}{2}(\text{sign}(b)+1)$ is the Heaviside step function.
	When evaluating the objective, $\ell_{\rm cal}(\thetav;\CD)$ in~\eqref{eq:km_emd}, $\hat{T}_n$ is either a sample from the model, $\hat{T}_n=g_{\thetav}(r_{\bs\psi}(\bs{x}),\bs{\epsilon})$, or an observed time $\hat{T}_n\sim p(t)$, for $\hat{S}_{\rm PKM}^{g_{\thetav}(r_{\bs\psi}(\bs{x}),\bs{\epsilon})}(t_i)$ or $\hat{S}_{\rm PKM}^{p(t)}(t_i)$, respectively. Expression $\hat{S}_{\rm PKM}$ represents the point-estimate-based formulation of the Kaplan Meier estimator, see  \citep{chapfuwa2019sfm}  for details.
	
	\subsection{Learning}
	For joint learning of all model parameters, $\{\bs{c}_k\}_{k=1}^K$, $\bs{\psi}$ and $\bs{\theta}$, we optimize both the latent representation and time-to-event (accuracy and calibration) objectives.
	The complete objective function for the proposed Survival Cluster Analysis (SCA) model is
	\begin{align}\label{eq:cost}
		\begin{aligned}
			\ell(\thetav,\bs{\psi},\{\bs{c}_k\}_{k=1}^K;\mathcal{D}) & = \ell_{\rm dp}(\bs{\psi},\{\bs{c}_k\}_{k=1}^K;\mathcal{D}) \\
			& + \lambda_2\ell_{\rm acc}(\thetav,\bs{\psi};\mathcal{D}) + \lambda_3\ell_{\rm cal}(\thetav,\bs{\psi};\CD) \,,
		\end{aligned}
	\end{align}
	where $\lambda_2,\lambda_3 > 0$ are hyper-parameters controlling the trade-off between accuracy and  calibration objectives, relative to the clustering objective in \eqref{eq:clust_obj}.
	For simplicity and comparability with SFM, we set $\lambda_2 = \lambda_3=1$.
	The objective in \eqref{eq:cost} is optimized using stochastic gradient descent on minibatches from $\mathcal{D}$.
	
	In practice, $\{\bs{c}_k\}$ is updated according to stochastic gradient descent by optimizing the KL objective \eqref{eq:clust_obj}, and is initialized with $K$-means after pretraining \eqref{eq:cost} without the clustering objective.
	During inference, we assign a new observation, $\bs{x}_\star$, to a cluster by first evaluating $q(u_\star=k|\bs{x}_\star)$ for $k=1,\ldots,K$, then obtaining a hard assignment according to  $u_\star = {\rm argmax}_{k} \ q(u_\star=k|\bs{x}_\star)$.
	
\begin{figure*}
		\centering
		\begin{tabular}{ccc}
			\includegraphics[width=0.33\linewidth,trim={0 0 0 15mm},clip]{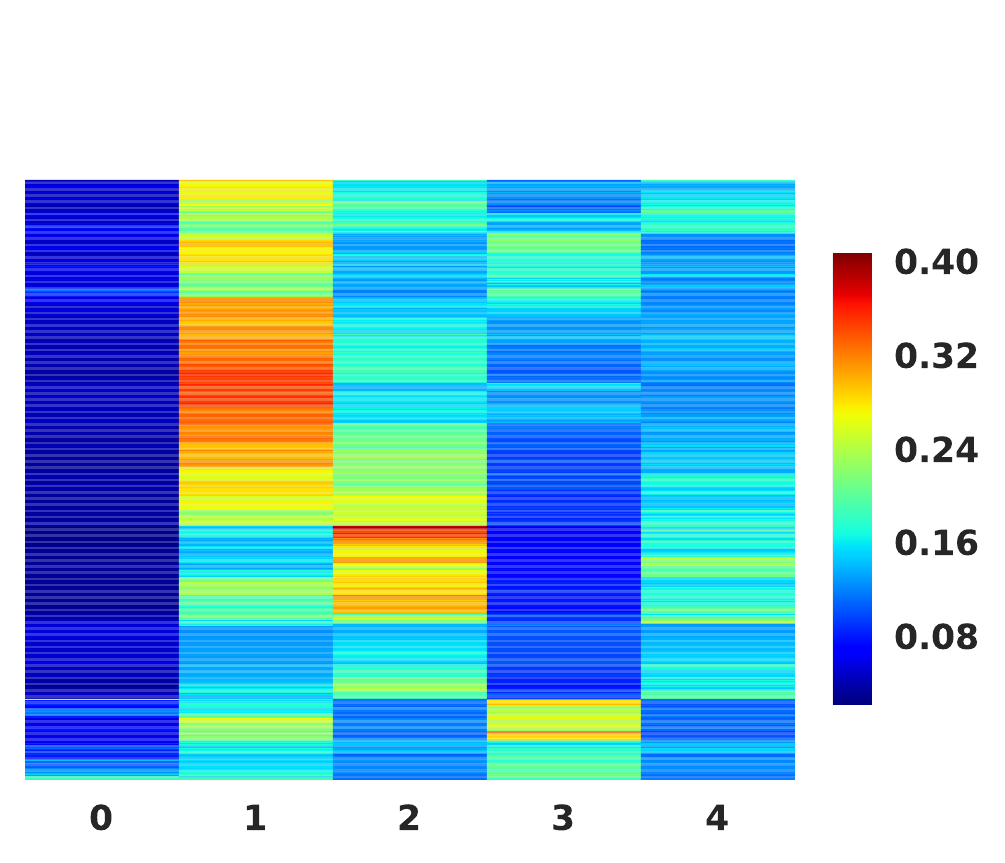} &
			\includegraphics[width=0.30\linewidth]{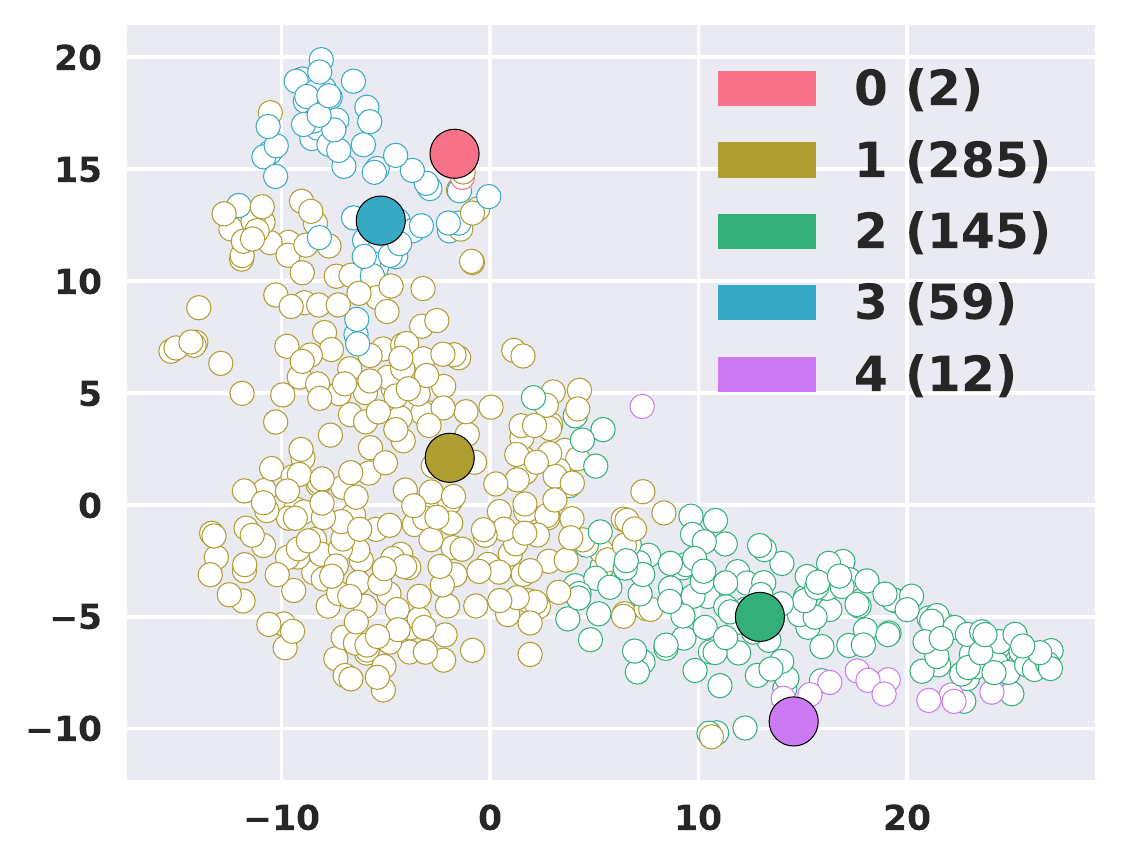} &
			\includegraphics[width=0.29\linewidth]{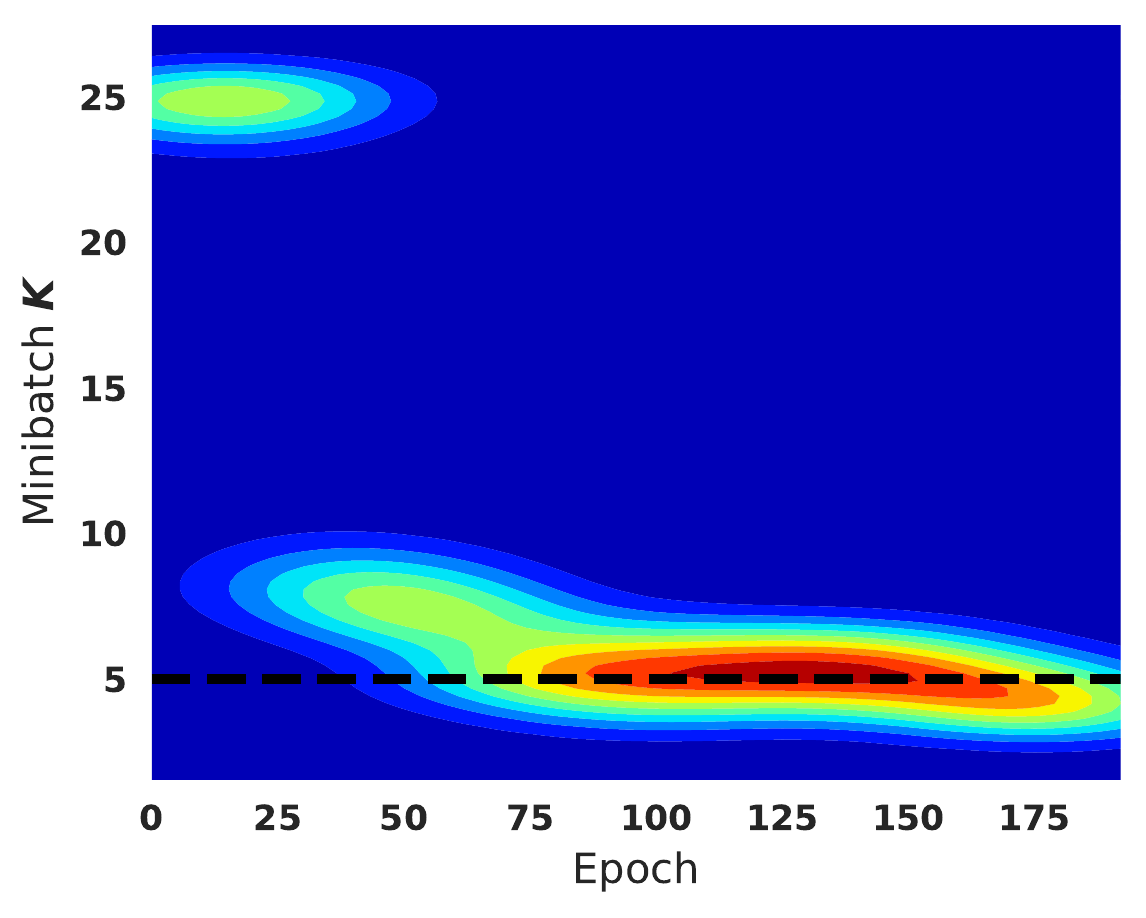}
			\\
			(a) Cluster assignments &
			(b) $t$-SNE plot  &
			(c) Inferred minibatch $K$  
		\end{tabular}		
		\vspace{-2mm}
		\caption{Inferred clusters on the testing set of {\sc sleep} dataset, with $K=25$ and $\gamma_{o}=3$ where: (a) corresponding individual probability distribution $q(\bs{\pi}|\bs{\xi},\{\bs{x}_1\}_{n=1}^M)$, are approximated according to \eqref{eq:qpi}, (b) joint $t$-SNE plot of centroids $\bs{c}_k$ with latent representation $\bs{z}$ and (c) density plot of inferred number of clusters $K$ during training.
		}
		\label{fg:clusters_seer}
		\vspace{-2mm}
	\end{figure*}

	The maximum number of mixture components $K$ is fixed during learning.
	However, provided that the KL divergence \eqref{eq:clust_obj} encourages mixture proportions to follow a \emph{stick-breaking process}, the {\em effective} number of mixture components, {\em i.e.}, those with non-empty observation assignments, will be smaller than $K$, thus effectively resulting in the model learning the number of mixture components.
	This is illustrated in Figure~\ref{fg:clusters_seer} and described below in the experiments.  The number of degrees of freedom, $\nu$ is a hyperparameter, set to 1 in our experiments.
	
	\section{EXPERIMENTS}\label{sc:exp}
	The comparisons presented below are made across a diverse range of six datasets, as summarized in Table~\ref{tb:data}.
	Refer to the Supplementary Material for all details concerning the experimental setup.
	Throughout the experiments, we set $K = 25$ and select $\gamma_o = \{2,3, 4, 8\}$ via grid search cross-validation from the training sets. TensorFlow code to replicate experiments can be found at \textcolor{blue}{\url{https://github.com/paidamoyo/survival_cluster_analysis}}.
	
	\paragraph{Datasets}
	Table~\ref{tb:data} shows the summary statistics of the six datasets considered.
	The datasets are diverse in number of observations $N$, varying amounts of categorical (cat) and continuous covariates $d$, proportions of non-censored events, missingness rates in the $N \times d$ covariate matrix, and time horizon $t_{\rm max}$ (measured in days, except for {\sc seer} which is measured in months). Following information-theoretic data processing inequality conclusions from \citep{miscouridou2018deep}, demonstrating insignificant performance change relative to pre-imputation, we impute continuous and categorical covariates with the median and mode,  respectively. In our experiments we do not convert time to a common scale but model it {as is}. 
	
	Publicly accessible datasets include:
	$i$) {\sc flchain}: a study of non-clonal serum immunoglobin free light chains effects on survival time \citep{dispenzieri2012use}.
	$ii$) {\sc support}: investigates the survival time  of seriously-ill hospitalized adults \citep{knaus1995support}.
	$iii$) {\sc seer}:  accessible from the Surveillance, Epidemiology, and End Results (SEER) Program.  The dataset is preprocessed according to the details described in  \citep{ries2007cancer}. 
	We restrict the dataset to a 10-year follow-up breast cancer subcohort. 
	
	The following datasets are available upon request: 
	$iv$) {\sc ehr}: a large study from the Duke University Health System centered around multiple inpatient visits due to comorbidities in patients with Type-2 diabetes \citep{chapfuwa2018adversarial}.
	$v$) {\sc sleep}: a subset of the Sleep Heart Health Study (SHHS) \citep{quan1997sleep}, a multi-center cohort study implemented by the National Heart Lung \& Blood Institute to determine the cardiovascular and other consequences of sleep-disordered breathing. We focus on the baseline clinical visit and aggregated demographics, medications and questionnaire data as covariates.
	$vi$) {\sc framingham}: a subset (Framingham Offspring) of the  longitudinal study of heart disease \citep{benjamin1994independent} dataset, initially for predicting 10-year risk for future coronary heart disease (CHD).
	
	\paragraph{Clustering Baselines} We consider the standard $K$-means and CoxPH based SSC-Bair \citep{bair2004semi} as strong clustering baselines for SCA.
	We provide quantitative evaluations in terms of the logrank score \citep{mantel1966evaluation}, and qualitative visualization of the clustering-based Kaplan-Meier sub-population survival curves.
	%
	%
	
	\paragraph{Time-to-Event Baselines} We compare SCA to the following time-to-event baselines: SFM \citep{chapfuwa2019sfm}, DATE \citep{chapfuwa2018adversarial}, S-CRPS \citep{avati2018countdown}, CoxPH \citep{cox1992regression}, MTLR \citep{yu2011learning} and DRAFT \citep{chapfuwa2018adversarial}.
	From these, SFM and DATE are key to our comparisons because we leverage components from their formulation into SCA; namely, the accuracy loss from DATE and the distribution matching loss from SFM.
	In that sense, we expect SCA to perform as good as SFM and DATE, but with the added benefit of producing clusters with distinct risk profiles.
	We present quantitative evaluations in terms of C-index, Calibration slope, Relative Absolute Error (RAE), and mean Coefficient of Variation (CoV).
	Details of these metrics are provided in the Supplementary Material.
	%
	%
	%
	
	\begin{table*}[t!]
		\centering
		\caption{\small Inferred cluster specific covariate information on the testing set for the {\sc framingham } dataset. The inferred cluster assignments are according to the corresponding individual probability distribution $q(\bs{\pi}|\bs{\xi},\{\bs{x}_1\}_{n=1}^M)$, approximated according to \eqref{eq:qpi}. Ranges in parentheses are 50\% empirical ranges over (median) test-set predictions for the continuous  and proportions for categorical covariates.}
		\begin{normalsize}
			\resizebox{0.98\linewidth}{!}{
				\begin{tabular}{lrrrrrrr}
					\toprule  
					Covariates & {\sc Cluster 0} & {\sc Cluster 1} & {\sc Cluster 2} & {\sc Cluster 3} & {\sc Cluster 4}
					& {\sc Cluster 5}
					& {\sc Cluster 6}\\
					\midrule
					Continous & & & & & & &\\
					\hdashline
					Age & $56_{(48,62)}$ & $50_{(43,58)}$ & $59_{(52,63)}$ & $55_{(48,61)}$ & $47_{(35,54)}$&
					$55_{(49,62)}$&
					$58_{(50,65)}$\\ 
					HDL (mg/dL) & $43_{(37,53)}$ & $52_{(44,63)}$ & $67_{(59,85)}$ & $54_{(45,66)}$ & $62_{(55,70)}$&
					$41_{(35,48)}$&
					$42_{(36,52)}$\\
					Total Cholesterol & $198_{(193,207)}$ & $176_{(168,183)}$ & $266_{(250,285)}$ & $220_{(207,236)}$ &
					$148_{(138,157)}$ &
					$251_{(235,275)}$ &
					$173_{(158,188)}$\\
					Systolic Blood Pressure & $126_{(117,137)}$ & $110_{(102,119)}$ & $141_{(130,153)}$ & $115_{(106,125)}$ &
					$110_{(102,117)}$&
					$126_{(115,139)}$&
					$132_{(120,147)}$\\
					\toprule
					Categorical & & & & &\\
					\hdashline
					Hypertension medication (Yes) & $25.5\%$ & $4.97\%$ & $40.1\%$ & $11.3\%$ & $1.1\%$&$41.6\%$&$41.0\%$\\
					Diabetic (Yes) & $6.9\%$ & $2.63\%$ & $3.0\%$ & $3.3\%$ & $0.0\%$& $20.8\%$& $16.7\%$\\
					Gender (Female) & $36.9\%$ & $82.5\%$ & $63.6\%$ & $69.6\%$ & $74.5\%$& $33.4\%$& $36.4\%$\\
					Current smoker (Yes) & $23.9\%$ & $14.6\%$ & $28.0\%$ & $16.6\%$ & $22.3\%$ & $45.6\%$ & $25.1\%$\\
					Race (Black) & $16.4\%$ & $3.5\%$ & $29.5\%$ & $1.5\%$ & $8.5\%$ & $21.7\%$ & $27.7\%$ \\
					Race (Chinese) & $4.2\%$ & $2.6\%$ & $0.0\%$ & $1.5\%$ & $2.1\%$ & $1.1\%$& $3.0\%$\\
					Race (Hispanic) & $5.0\%$ & $2.3\%$ & $2.3\%$ & $2.3\%$ & $1.0\%$ & $4.0\%$& $5.0\%$\\
					Race (White) & $74.4\%$ & $91.5\%$ & $68.2\%$ & $85.7\%$ & $88.3\%$ & $73.2\%$& $64.2\%$\\
					\bottomrule
					\label{tb:pooled_cov}
				\end{tabular}
			}
		\end{normalsize}
		\vspace{-5mm}
	\end{table*}	
	
	\subsection{Qualitative Results}
	Figure~\ref{fg:clusters_seer} shows for the {\sc sleep } dataset $a)$ estimated individualized cluster assignment probability distributions (rows) evaluated according to \eqref{eq:qpi}; $b)$ $t$-SNE plots of the model inferred centroids, $\bs{c}_k$, as well as the individual latent representation $\bs{z} = r_{\bs\psi}(\bs{x})$; and $c)$ density plot of the inferred number of (non-empty) clusters $K$ during training.
	See the Supplementary Material for similar figures for all the other datasets, where we also include corresponding Kaplan-Meier curves, as in Figure~\ref{fig:clustering}.
	
	\begin{figure}[t!]
		\centering
		\includegraphics[width=0.95\linewidth]{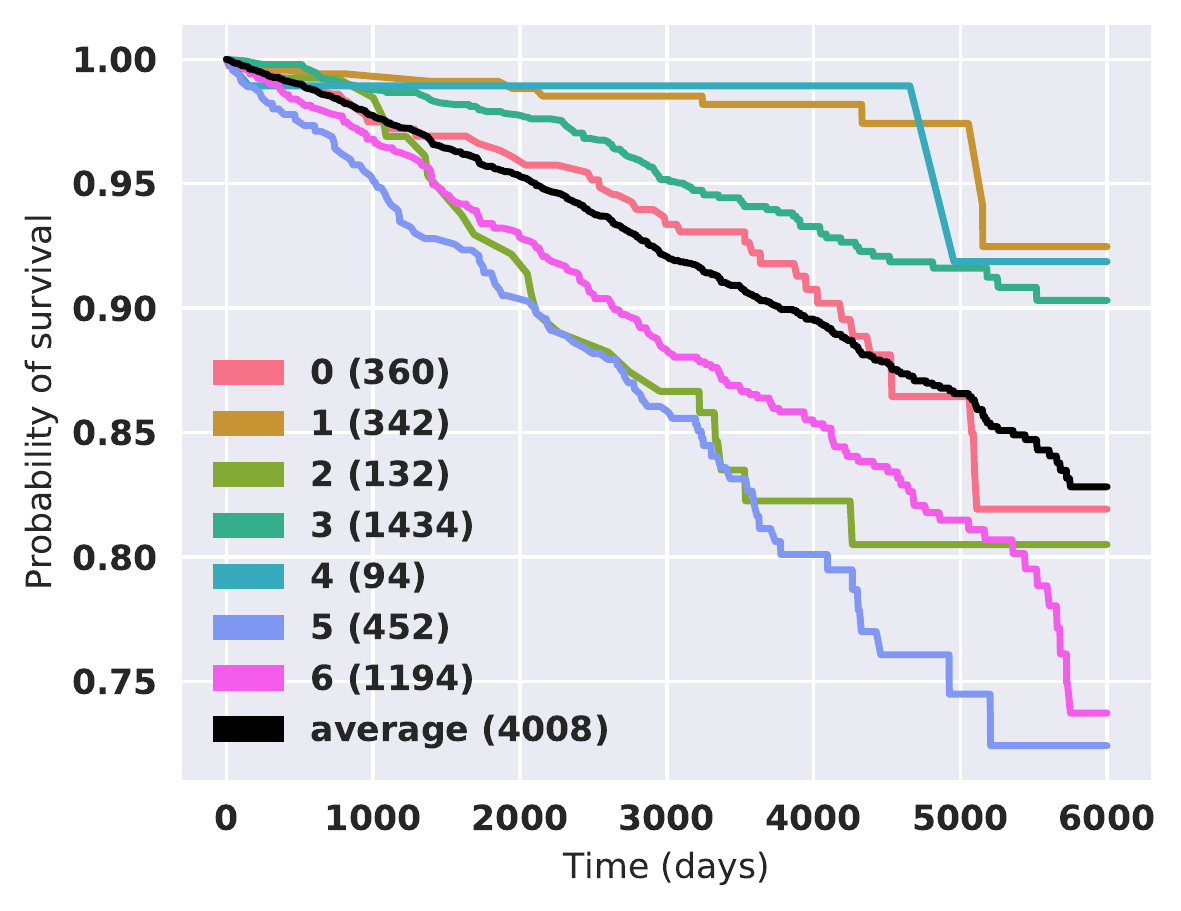}
		\vspace{-2mm}
		\caption{Inferred Cluster specific Kaplan-Meir Curves on the testing set of {\sc Framingham} dataset, with $K=25$ and $\gamma_{o}=8$. The inferred clusters assignment is according to the corresponding individual probability distribution $q(\bs{\pi}|\bs{\xi},\{\bs{x}_1\}_{n=1}^M)$, approximated according to \eqref{eq:qpi}.}
		\label{fig:pooled_km}
	\end{figure}
	
	Interestingly, the cluster-specific covariate statistics for the {\sc Framingham} dataset, which has the least number of covariates, are shown in Table~\ref{tb:pooled_cov} and are consistent with findings from the Framingham Heart Study \citep{benjamin1994independent}, which identified high blood cholesterol and high blood pressure as major risk factors for cardiovascular disease.
	
	We obtain the cluster specific Kaplan-Meir curves illustrated in Figure~\ref{fig:pooled_km} with corresponding cluster specific covariate information shown in Table~\ref{tb:pooled_cov}.
	The inferred individual cluster assignment is obtained according to the individual probability distribution $q(\bs{\pi}|\bs{\xi},\{\bs{x}_1\}_{n=1}^M)$, approximated according to \eqref{eq:qpi}. We consider curves above the population average low-risk while the curves below to be high-risk.
	
	Therefore, our model identifies three high-risk clusters, indexed by 2, 5, 6: 
	$i)$ cluster 6 and cluster 5 have similar statistics, as they both consists disproportionately of diabetic individuals on hypertension medication with elevated total cholesterol (normal is below 200), high systolic blood pressure (normal is below 120), and noticeably low HDL (normal is greater than 60);
	$ii)$ cluster 2, is also driven by age, which is expected, where about 40\% of the population is on hypertension medication, and with the worst systolic blood pressure and cholesterol compared to other clusters; 
	$iii)$ lower-risk clusters 1, 3, 4 are mostly comprised of females with normal levels of HDL, total cholesterol and systolic blood pressures;
	$iv)$ cluster 0 represents the average statistics of the Framingham dataset, thus the survival curves directly follows the empirical population survival.
	Finally, note that the three high-risk clusters (2, 5 and 6) have a substantial over-representation of African Americans, known to have an increased risk for cardiovascular disease \citep{benjamin1994independent}. See the Supplementary Material for additional inferred cluster specific Kaplan-Meir curves on all datasets.

	We demonstrate that by jointly learning clustering with respect to both the covariates $\xv$ and predicted time-to-event $t$, our model SCA can identify high-, medium- and low-risk individuals 
	, which is essential for clinical decision making.
	During inference, both the risk profile and individualized time-to-event can provide a comprehensive prediction mechanism for identifying cluster-based risk factors, cluster-based risk profiles and individualized time predictions.
	Further, the advantage of matching the empirical mixture distribution with a (truncated) DP yields sparse predictions of cluster assignment probabilities, $q(\bs{\pi}|\bs{\xi},\{\bs{x}_1\}_{n=1}^M)$, manifested as high confidence cluster assignments illustrated as a heatmap Figure~\ref{fg:clusters_seer}(a).
	
\begin{figure}[t!]
		\centering
		\includegraphics[width=0.95\linewidth]{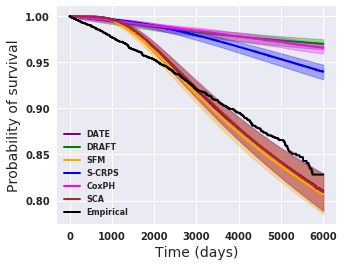}\\
		\small (a) {\sc Framingham}
		%
		%
		\includegraphics[width=0.95\linewidth, height=5.4cm]{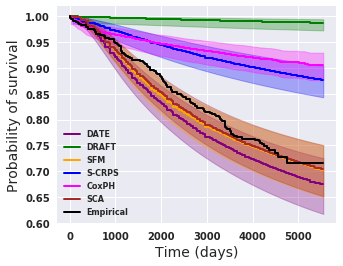}\\
		\small (b) {\sc sleep}
		%
		\caption{\small Survival function estimates for (a) {\sc Framingham} and (b) {\sc sleep} data. Ground truth (Empirical) is compared to predictions from six models (DATE, DRAFT, SCA (our proposed model), SFM, S-CRPS and CoxPH). Error bars (shaded area) are calculated according to the Greenwood's formula \citep{greenwood1926report}.}
		\label{fg:surv_funct}
	\end{figure}

	\paragraph{Calibration Curves}
	We visually compare calibration curves from DATE, DRAFT, SCA, SFM, S-CRPS and CoxPH. Figure \ref{fg:surv_funct} shows the estimated \emph{populations-based} model survival functions according to \citep{chapfuwa2019sfm} and empirical Kaplan-Meier for the {\sc Framingham} and {\sc sleep} datasets. Error bars (shaded area) are calculated according to the Greenwood's formula \citep{greenwood1926report}. For all datasets, SCA- and SFM-estimated population survival functions closely match the empirical ground truth survival function, which is consistent with the high calibration slopes results in Table ~\ref{table:quant_results}.
	See the Supplementary Material for additional calibration and survival function results on all datasets.
	%
	
	\subsection{Quantitative Results}
	\begin{table*}
		\centering
		\caption{Calibration slope and RAE metrics on test data.}
		\vspace{-2mm}
		\resizebox{0.85\linewidth}{!}{
			\label{table:quant_results}
			\begin{tabular}{lrrrrrr}
				& {\sc ehr} & {\sc flchain} & {\sc support} & {\sc seer}  & {\sc sleep} &  {\sc framingham}\\
				\toprule
				Calibration slope & & & & && \\
				\hdashline
				DATE  &0.7537  &0.9668 &0.9068 & 0.9161& 0.9454  & 0.7737\\
				DRAFT  &3.2138  & 5.4183 & 2.9640 &2.0763&  25.2855 &  5.7345\\
				S-CRPS &1.6246 & 1.9662& 1.1795&1.1613& 2.5746 & 2.6114\\
				CoxPH &2.5543&1.9116&1.3909&1.4358& 3.8278 & 4.9945 \\
				MTLR  &2.1957 & 1.9449 & 1.2017  & 1.2476 & 2.4792 & 5.4878\\
				SFM   &   0.7734 & 0.9807  &  0.9405 & \textbf{0.9540} & 1.0235  &0.7626 \\
				SCA (proposed) & \textbf{0.8006} &\textbf{ 0.9900} & \textbf{1.0086}& 0.9290& \textbf{1.0223} & \textbf{0.8044}\\
				\toprule
				RAE (non-censored)  & & & & & &\\
				\hdashline
				DATE   & \textbf{0.6107} & 0.5222& 0.6691& 0.5289&0.5224 & 0.5122\\
				DRAFT   & 0.7099 & 0.6399& 0.7109& 0.6097&  0.7465& 0.6697\\
				S-CRPS  & 0.7240 & 0.6378& \textbf{0.4851}& 0.5323&0.7330 & 0.8369 \\
				CoxPH   & -&- & -& -&-&- \\
				MTLR  & -&- & -& -&-&- \\
				SFM   &0.6146 & \textbf{0.5111}& 0.6398& 0.5294& \textbf{0.5162}&  \textbf{0.5074}\\
				SCA (proposed)  & 0.6186 &0.5134 &0.6295 & \textbf{0.5193}& 0.5424& \textbf{0.5074}\\		
				\bottomrule
			\end{tabular}
		}
	\end{table*}
	
	\begin{table*}
		\caption{Logrank score and standard errors in parentheses. The best performing $K$-means and SSC-Bair models were selected from the set $K= \{2, 3, 4, 5, 6\}$.}
		\centering
		\vspace{-2mm}
		\resizebox{0.85\linewidth}{!}{
			\begin{tabular}{lrrrrrr}
				& {\sc ehr} & {\sc flchain} & {\sc support} & {\sc seer}  & {\sc sleep} &  {\sc framingham}\\
				\toprule
				SCA &-&\textbf{278.49 (0.0)} & 496.50 (0.0)&\textbf{4803.59 (0.0)}& \textbf{63.74 (0.0)}& 123.31 (0.0)\\
				SSC-Bair &409.93 (0.0)&4.27 (0.37)& \textbf{1204}.14 (0.0)&4084.78 (0.0) &20.37  (0.0)& \textbf{125.87} (0.0)\\
				K-means & 417.00 (0.0)&5.33 (0.38)&99.06 (0.0)&3985.67 (0.0) &21.83 (0.0)&88.87 (0.0)\\
				\bottomrule
			\end{tabular}
		}
		\label{tb:clust_results}
	\end{table*}
	
	Below we describe performance metrics across all datasets and models.
	Specifically, calibration slope, mean CoV (coefficient of variation), C-index \citep{harrell1984regression} and Relative Absolute Error \citep[RAE,][]{yu2011learning} provide a comprehensive evaluation, as they offer insights into consistency of time-to-event predictions, concentration of predicted distributions, pairwise ranking consistency, and accuracy of event time predictions, respectively. 
	The results demonstrate that by jointly modeling the time-to-event and cluster assignments we obtain a better calibrated model, that is competitive in C-index, concentrated and accuracy of predictions.
	Table~\ref{table:quant_results} shows the calibration slopes and RAE across all datasets and models.
	See the Supplementary Material for detailed RAE, mean CoV and C-index results.
	
	%
	%
	
	Table \ref{tb:clust_results} presents the clustering performances of the best performing $K$-means, SSC-Bair and SCA algorithms, measured in terms of the logrank score \citep{mantel1966evaluation}, for SSC-Bair and $K$-means we selected the best performing model from the set $K=\{2, 3, 4, 5, 6\}$.
	
	\paragraph{Calibration Slope}
	For calibration we use the framework developed in SFM to evaluate the models \citep{chapfuwa2019sfm}. An ideal calibration slope is 1, while a slope $<1$ and slope $>1$ 
	indicates whether the model tends to underestimate or overestimate risk, respectively. The clustering objective in SCA augments the calibration objective we borrow from SFM, thus improving the calibration even for \emph{non-iid} observations, such as {\sc Framingham} and {\sc ehr}, which are considered poorly calibrated, as illustrated in SFM.
	Given that SCA leverages the calibration objective of SFM, it is not surprising that both SCA and SFM are competitive, followed by DATE, S-CRPS, MTLR, CoxPH and lastly DRAFT. See Supplementary Material for qualitative calibration plots.
	
	\paragraph{Relative Absolute Error (RAE)} We compute RAE for both censored and non-censored events. In Table \ref{table:quant_results} we present the RAE for non-censored event times ($l_n=1$) for models that predict absolute event times, thus excluding scoring based models (CoxPH and MTLR). The results demonstrate that DATE, SFM and SCA (nonparametric) methods outperform DRAFT and S-CRPS (parametric) methods, which is expected since they all use a similar accuracy-aware objective function. For censored events ($l_n=0$), RAE provides the lower bound error given the censored time provides tail information of $p(t|\xv)$; parametric methods (DRAFT and S-CRPS) have small advantage over nonparametric methods (SFM, DATE and SCA). See the Supplementary Material for additional results on censored event times.
	
	\paragraph{Concordance Index (C-index)}
	C-index is a ranking metric that does not account for uncertainty in time-to-event predictions.
	Therefore to evaluate the time-to-event models (except CoxPH) in terms of C-index, we use point summaries of the individualized time-to-event distributions, specifically, $\hat{t}={\rm median}(\{t_{ns}\}_{s=1}^{200})$, where $t_{ns}$ is a sample from the trained model, $t_{ns} =g_{\thetav}(r_{\bs\psi}(\xv_{n}),\epsilonv_s)$ on the test set.
	Apart from the small covariates, the very low event rate {\sc Framingham} dataset and the small high event rate {\sc support} dataset, none of the models have a clear advantage on the C-index metric.
	This is not surprising because C-index with very low event rate is heavily influenced by the censored observations. Note, for MTLR, although we can compute the C-index at prespecified thresholds, we are unable to compute a global C-index.
	
	\paragraph{Coefficient of Variation (CoV) }
	Models that characterize the event time density function $f(t|\xv)$ result in uncertainty-aware time-to-event predictions.
	In practice, it is highly desirable for a model to generate concentrated time-to-event predictions.
	The CoV (coefficient of variation) measures the dispersion in a distribution; a $\text{Cov} > 1$ indicates high variance, while $\text{CoV} < 1$ indicates low variance distributions. Cov results provided in the Supplementary Material demonstrate that 
	DATE, SCA and SFM are consistently low-variance distributions, followed by S-CRPS and lastly DRAFT.
	We cannot compute CoV for both MTLR and CoxPH. CoxPH estimates risk score, and therefore cannot be evaluated on CoV.  MTLR does not specify the conditional hazards, $ h(t|\xv)$, and thus we cannot recover $f(t|\xv) = S(t|\xv)h(t|\xv)$.
	
	\paragraph{Logrank Score} 
	The logrank score is a nonparametric statistic that evaluates the similarity between a pair of survival functions, yielding high values for curves that are highly unlikely to be similar \citep{mantel1966evaluation}. Further, the logrank statistic is especially powerful for measuring differences between survival functions that follow the Cox proportional hazard assumption, $i.e$, the survival functions do not cross. For $K$ clusters, we compute ${K \choose 2} $ pairwise comparisons. Table~\ref{tb:clust_results} demonstrates that our proposed SCA is the best performing method, followed by SS-Bair and lastly $K$-means. Interestingly, SCA is unable to recover any clustering structure from the {\sc ehr} dataset, as it is a homogeneous population of Type-2 diabetes subjects, whereas $K$-means and SSC-Bair are always able to produce clusters (which may be misleading for homogeneous datasets). This supports the need to account for survival information when clustering survival datasets, as both SCA and SSC-Bair incorporate time information in their clustering approaches. 
	
	\section{CONCLUSIONS}
	We have developed the first time-to-event model for inferring individualized risk-based cluster assignments, while jointly predicting the time-to-event.
	Leveraging a Bayesian nonparametric stick-breaking representation of the Dirichlet Process, we have presented a method for learning a clustering structure in a latent representation, for which the number of clusters is unknown.
	We have demonstrated the need to account for time information when clustering survival datasets.
	Our model identifies interpretable and phenotyopically heterogeneous subpopulations, which are critical in a clinical setting for identifying subjects with diverse risk profiles.
	Extensive experiments demonstrate that the joint modeling approach yields substantial performance gains in calibration and logrank scores, while remaining competitive in preserving pairwise ordering, predicting concentrated and accurate distributions.
	In the future, we plan to extend this work to account for locally-consistent, calibrated and accurate predictions within identified subpopulations.
	
	\section*{Acknowledgments}
	The authors would like to thank the anonymous reviewers for their insightful comments.
	This work was supported by NIH/NIBIB R01-EB025020.
	
	
	\bibliographystyle{ACM-Reference-Format}
	\bibliography{sca}

\end{document}


\appendix
	\title{ Supplementary Material for ``Survival Cluster Analysis"}

	\maketitle
	
	\thispagestyle{empty}

	\section{Notation}
	Refer to Table \ref{tb:notation} for summary descriptions of notation used in the SCA formulation.

	\section{ Experimental Setup}
	In all experiments, SCA, SFM, DATE, DRAFT  and S-CRPS are specified in terms of two-layer MLPs of $50$ hidden units with Rectified Linear Unit (ReLU) activation functions, batch normalization \citep{ioffe2015batch} and apply dropout of $p=0.2$ on all layers. We set the minibatch size to $M=350$ and use the Adam \citep{kingma2014adam} optimizer with the following hyperparameters: learning rate $3 \times 10 ^{-4}$, first moment $0.9$, second moment $0.99$, and epsilon $1 \times 10 ^{-8}$.
	We initialize all the network weights according to  \textit{Xavier} \citep{glorot2010understanding}.  SFM and DATE inject noise in all layers, see \cite{chapfuwa2018adversarial} for more details; while SCA injects noise only in the last layer.
	Datasets are split into training, validation and test sets as 80\%, 10\% and 10\% partitions, respectively, stratified by non-censored event proportion. The validation set is used for early stopping and learning model hyperparameters. All models are trained using one NVIDIA P100 GPU with 16GB memory.
	
	\section{C-index, mean CoV and RAE Results}
	See Table \ref{tb:ci_rae_cov} for additional quantitative evaluations on C-index, mean CoV and RAE.

	\section{Calibration and Survival Function Results}
\begin{figure}[ht!]
		\centering
		\includegraphics[width=0.49\linewidth]{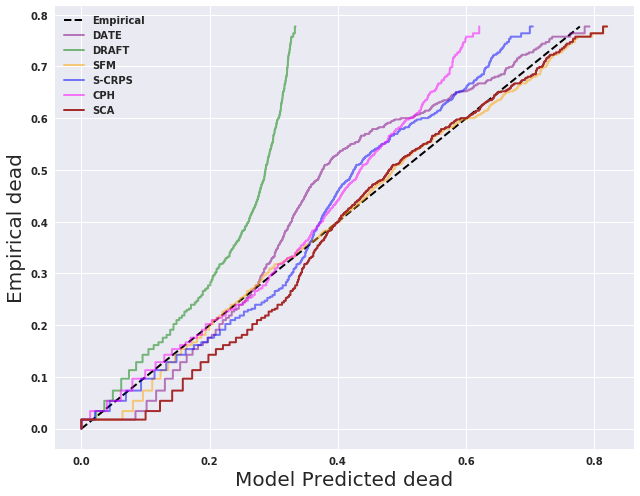}
		\includegraphics[width=0.49\linewidth]{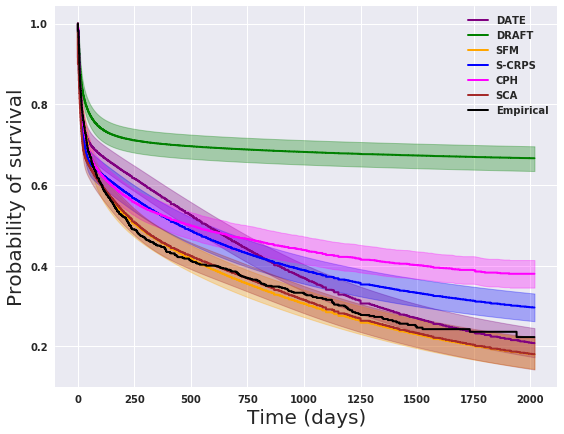}
		\caption{Calibration (left) and Survival function estimates (right) for {\sc support} data. Ground truth (Empirical) is compared to predictions from six models (DATE, DRAFT, SCA (our proposed model), SFM, S-CRPS and CoxPH).}
		\label{fg:support_calib}
	\end{figure}

\begin{figure}[ht!]
		\centering
		\includegraphics[width=0.49\linewidth]{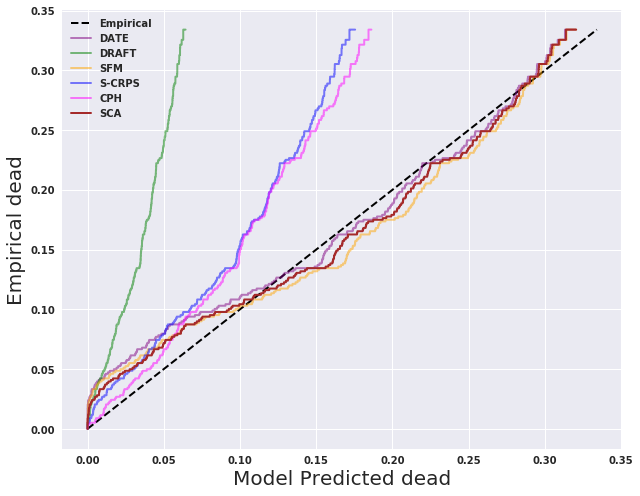}
		\includegraphics[width=0.49\linewidth]{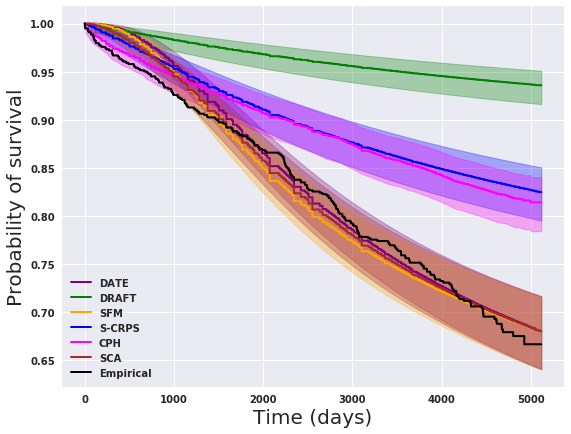}
		\caption{Calibration(left) and Survival function estimates (right) for {\sc flchain} data. Ground truth (Empirical) is compared to predictions from six models (DATE, DRAFT, SCA (our proposed model), SFM, S-CRPS and CPH).}
		\label{fg:flchain_calib}
	\end{figure}

\begin{figure}[ht!]
		\centering
		\includegraphics[width=0.49\linewidth]{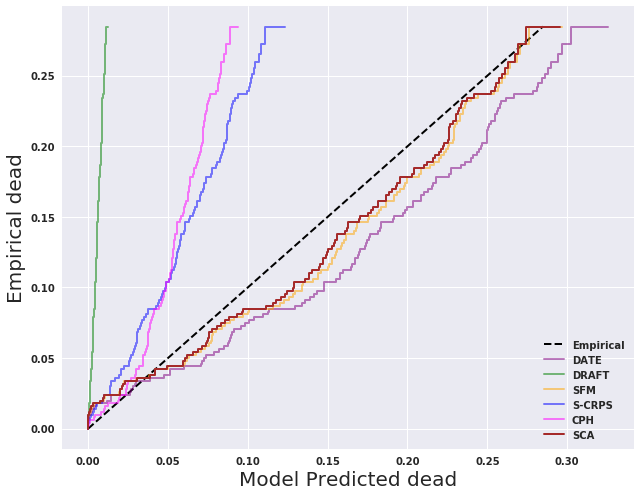}
		\includegraphics[width=0.49\linewidth]{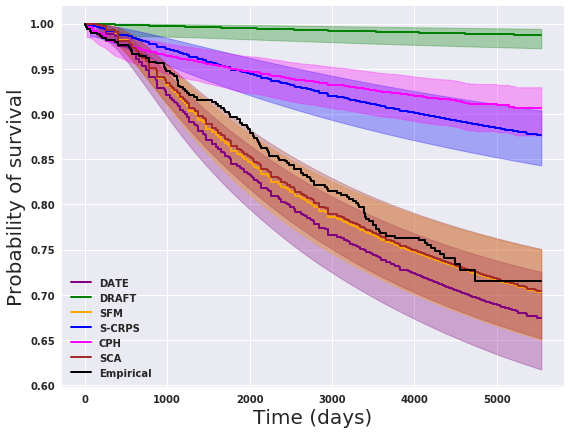}
		\caption{Calibration (left) and Survival function estimates (right) for {\sc sleep} data. Ground truth (Empirical) is compared to predictions from six models (DATE, DRAFT, SCA (our proposed model), SFM, S-CRPS and CPH).}
		\label{fg:sleep_calib}
		\vspace{2mm}
	\end{figure}

\begin{figure}[ht!]
		\centering
		\includegraphics[width=0.49\linewidth]{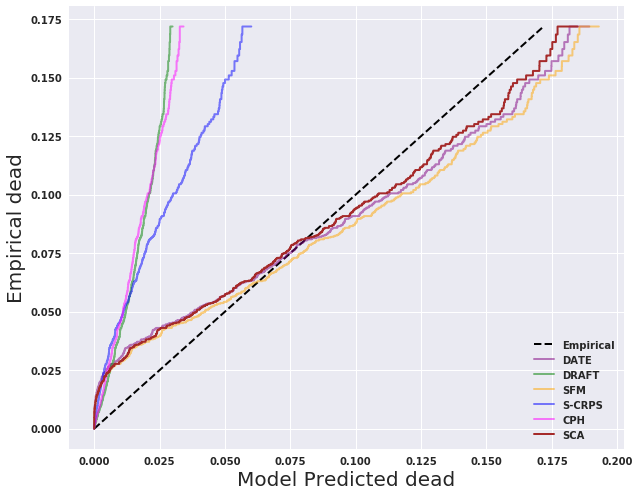}
		\includegraphics[width=0.49\linewidth]{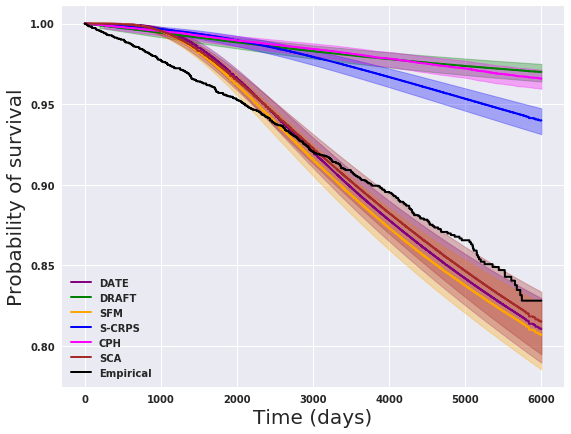}
		\caption{Calibration(left) and Survival function estimates (right) for {\sc framingham} data. Ground truth (Empirical) is compared to predictions from six models (DATE, DRAFT, SCA (our proposed model), SFM, S-CRPS and CPH).}
		\label{fg:pooled_calib}
		\vspace{2mm}
	\end{figure}

\begin{figure}[ht!]
		\centering
		\includegraphics[width=0.49\linewidth]{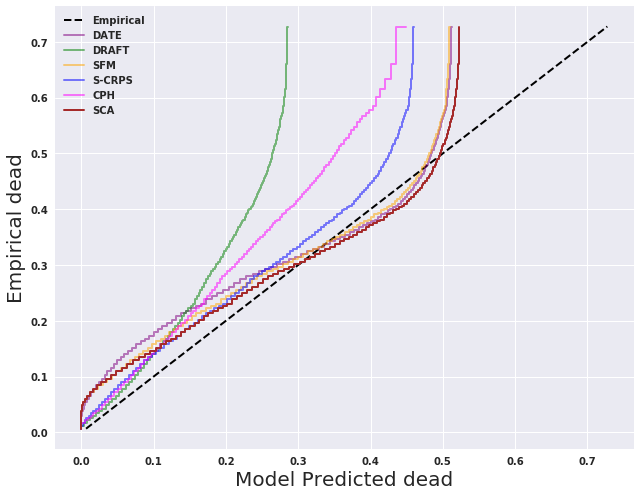}
		\includegraphics[width=0.49\linewidth]{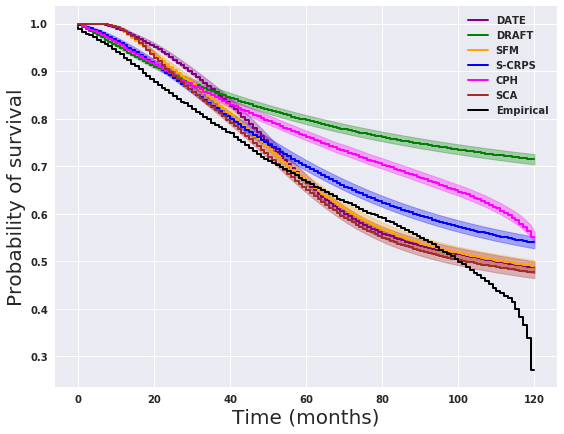}
		\caption{\small Calibration(left) and Survival function estimates (right) for {\sc seer} data. Ground truth (Empirical) is compared to predictions from six models (DATE, DRAFT, SCA (our proposed model), SFM, S-CRPS and CPH).}
		\label{fg:seer_calib}
		\vspace{2mm}
	\end{figure}

\begin{figure}[ht!]
		\centering
		\includegraphics[width=0.49\linewidth]{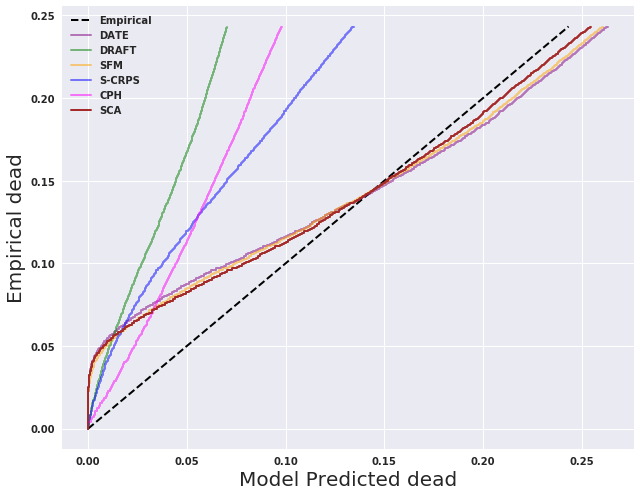}
		\includegraphics[width=0.49\linewidth]{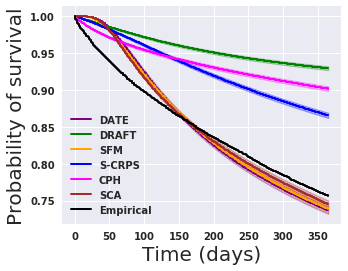}
		\caption{\small Calibration(left) and Survival function estimates (right) for {\sc ehr} data. Ground truth (Empirical) is compared to predictions from six models (DATE, DRAFT, SCA (our proposed model), SFM, S-CRPS and CPH).}
		\label{fg:sanofi_calib}
		\vspace{2mm}
	\end{figure}

	The model calibration and survival plots for datasets {\sc support}, {\sc flchain}, {\sc sleep}, {\sc seer}, {\sc framingham}, and {\sc ehr} are shown in Figures \ref{fg:support_calib} - \ref{fg:sanofi_calib}.

	\section{Latent-Space Representation Results}
\begin{figure*}[ht!]
		\centering
		\begin{tabular}{ccc}
			\includegraphics[width=0.32\linewidth,  height=4cm]{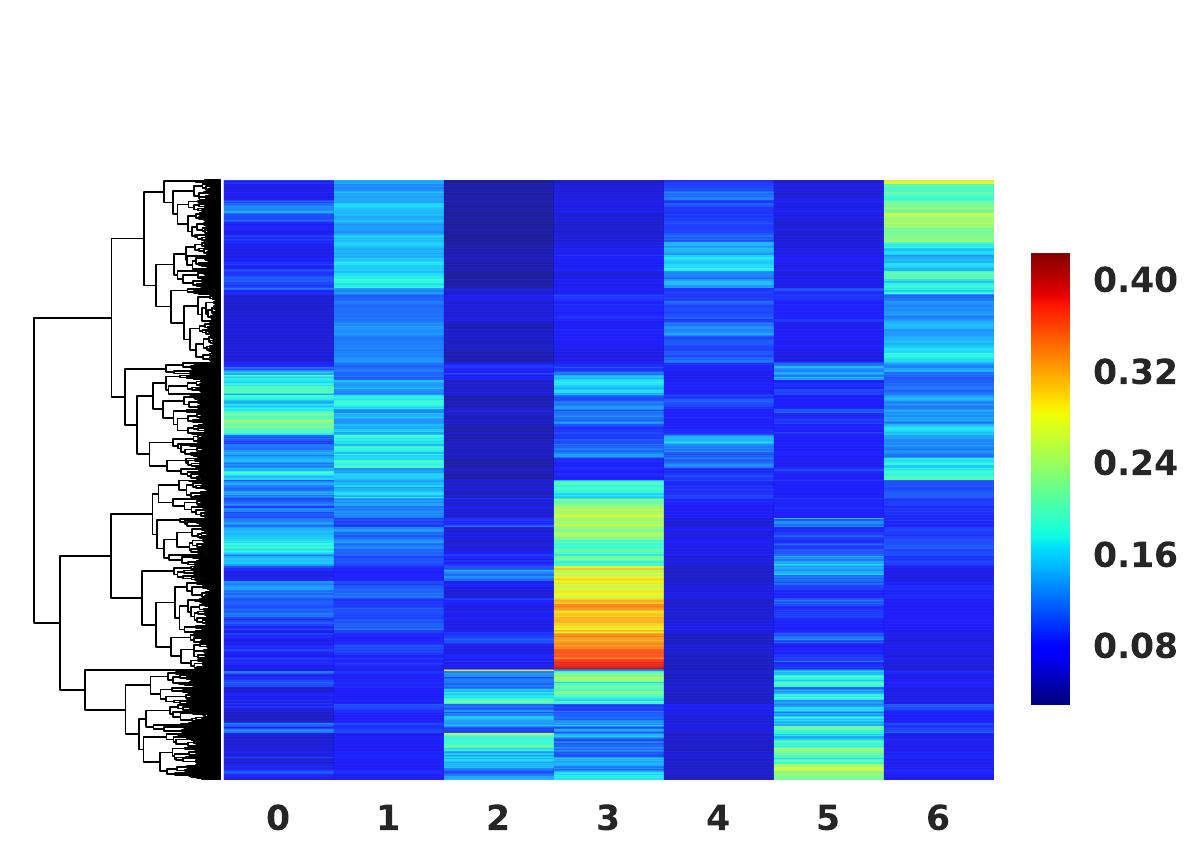}	&
			\includegraphics[width=0.32\linewidth,  height=4cm]{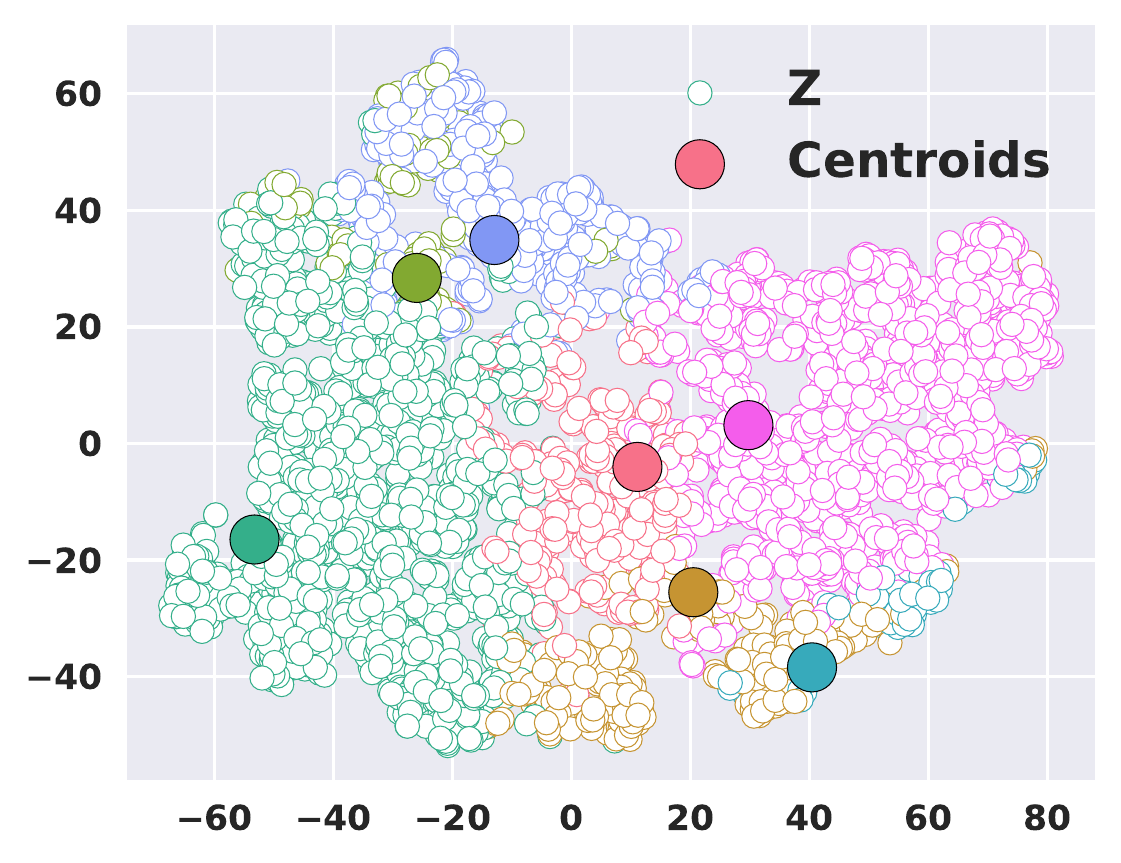}	&
			\includegraphics[width=0.32\linewidth, height=4cm]{sca_figures/pooled/pooled_learning_curve.pdf}
			\\
			(a) Cluster assignments &
			(b) t-SNE plot  &
			(c)  Cluster specific Kaplan-Meir Curves 
		\end{tabular}		
		\caption{Inferred clusters on the testing set of {\sc framingham} dataset, with $K=25$ and $\gamma_{o}=8$ with corresponding individual probability distribution $q(\bs{\pi} | \CD)$.
		}
		\label{fg:pooled}
	\end{figure*}

\begin{figure*}[ht!]
		\centering
		\begin{tabular}{ccc}
			\includegraphics[width=0.32\linewidth,  height=4cm]{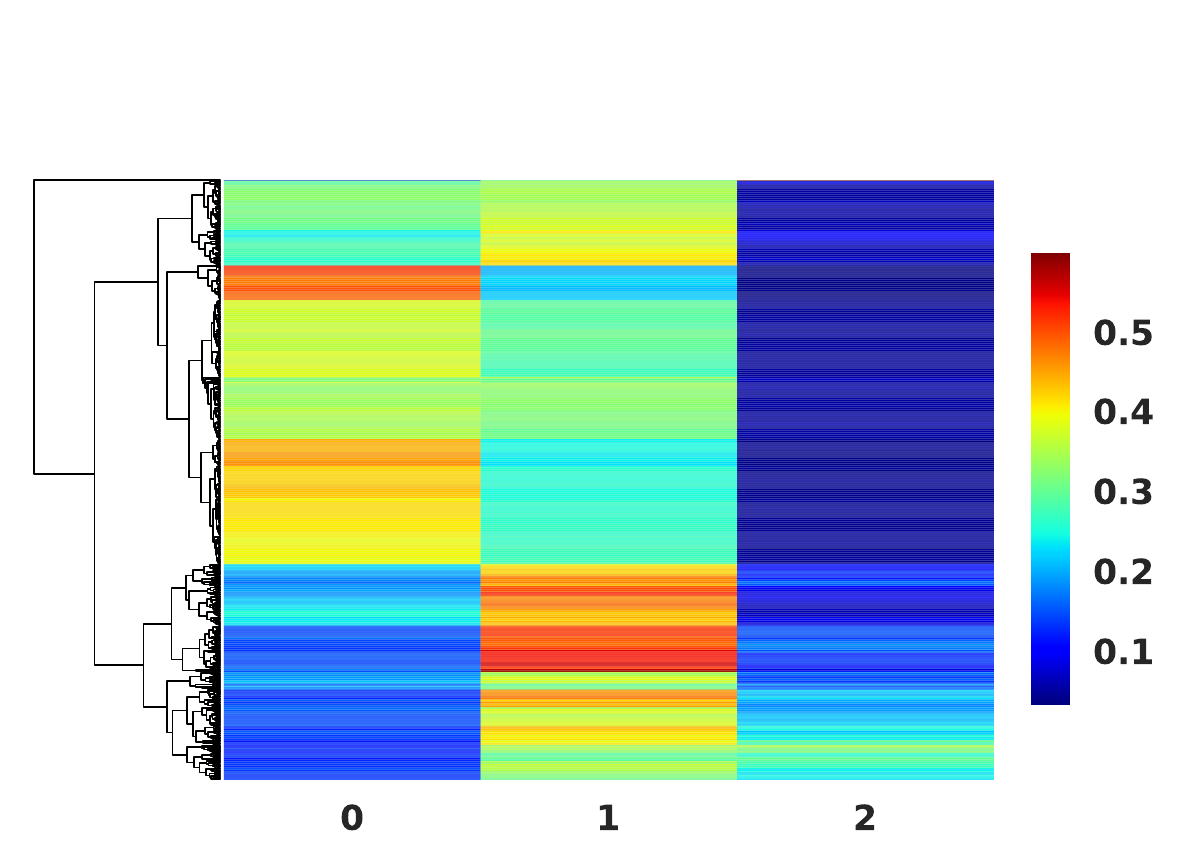}	&
			\includegraphics[width=0.32\linewidth,  height=4cm]{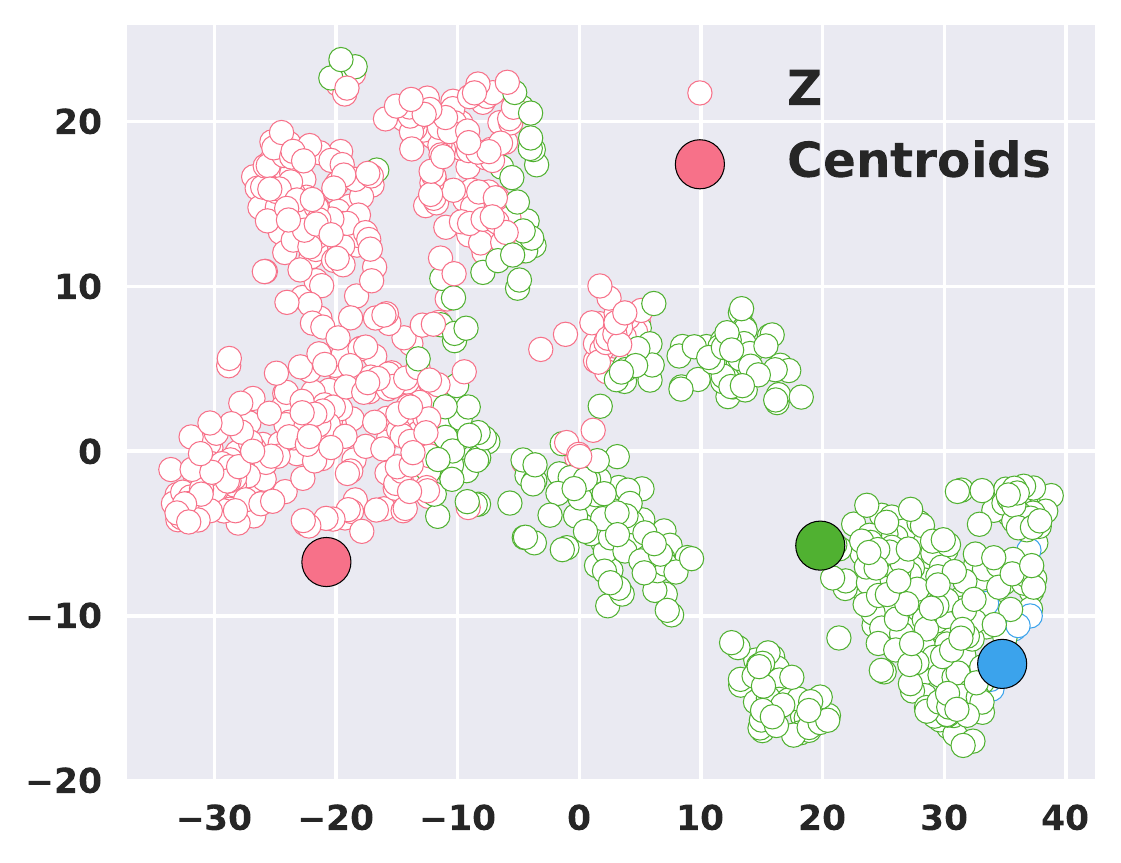}	&
			\includegraphics[width=0.32\linewidth, height=4cm]{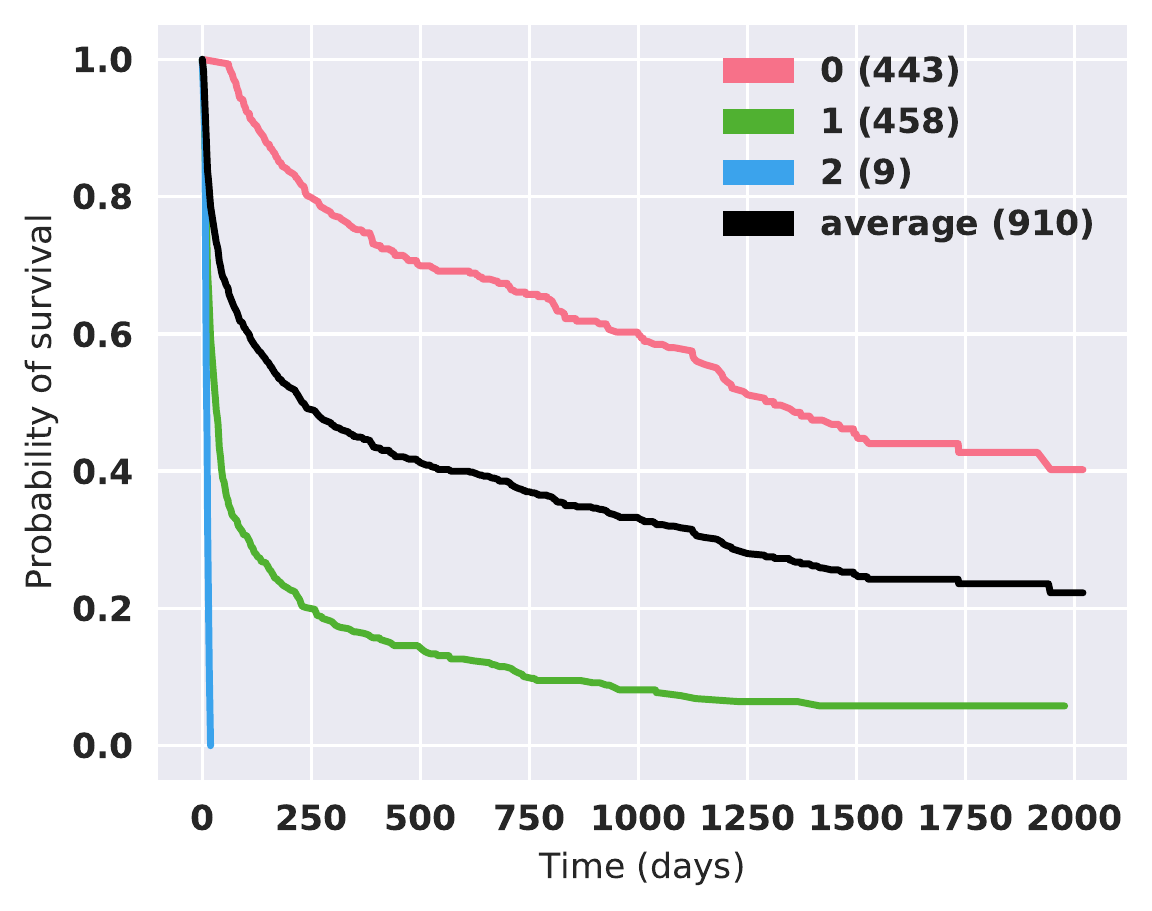}
			\\
			(a) Cluster assignments &
			(b) t-SNE plot  &
			(c)  Cluster specific Kaplan-Meir Curves 
		\end{tabular}		
		\caption{Inferred clusters on the testing set of {\sc support} dataset, with $K=25$ and $\gamma_{o}=2$ with corresponding individual probability distribution $q(\bs{\pi} | \CD)$.
		}
		\label{fg:support}
	\end{figure*}

\begin{figure*}[ht!]
		\centering
		\begin{tabular}{ccc}
			\includegraphics[width=0.32\linewidth,  height=4cm]{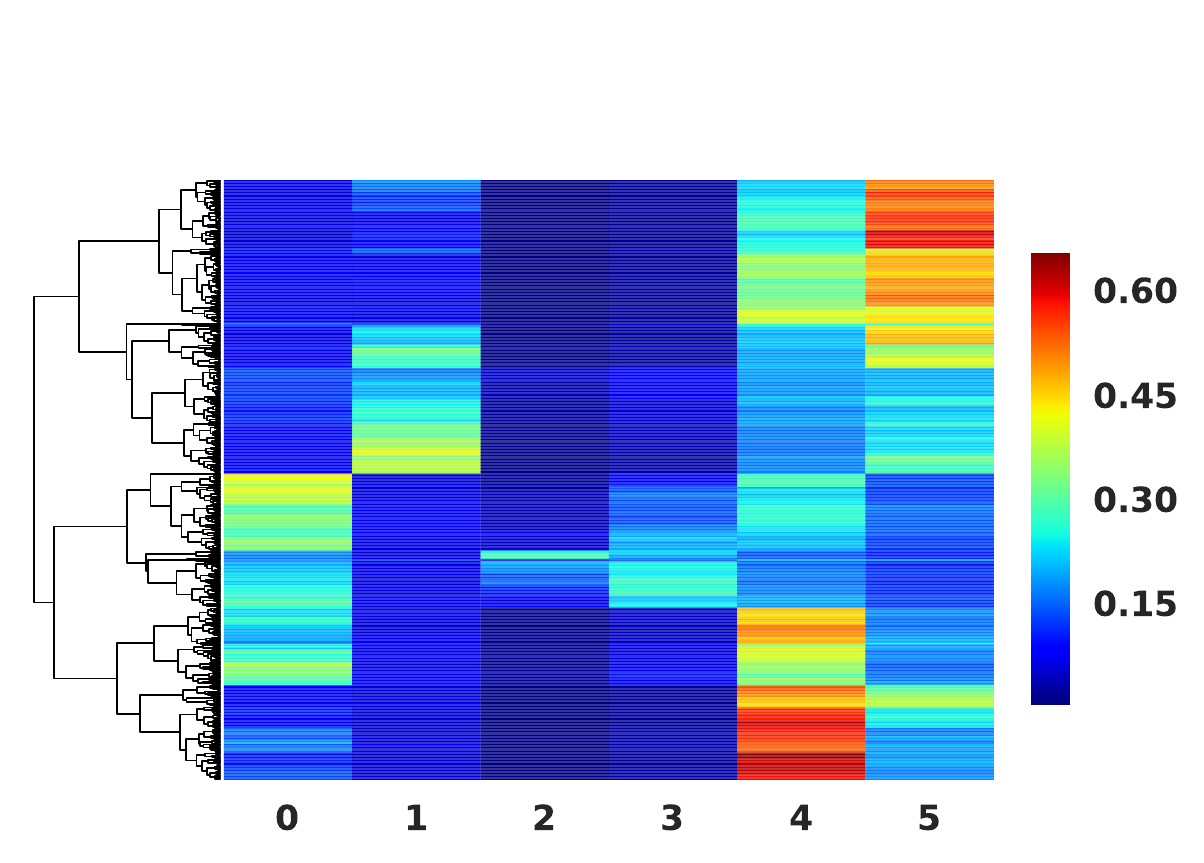}	&
			\includegraphics[width=0.32\linewidth,  height=4cm]{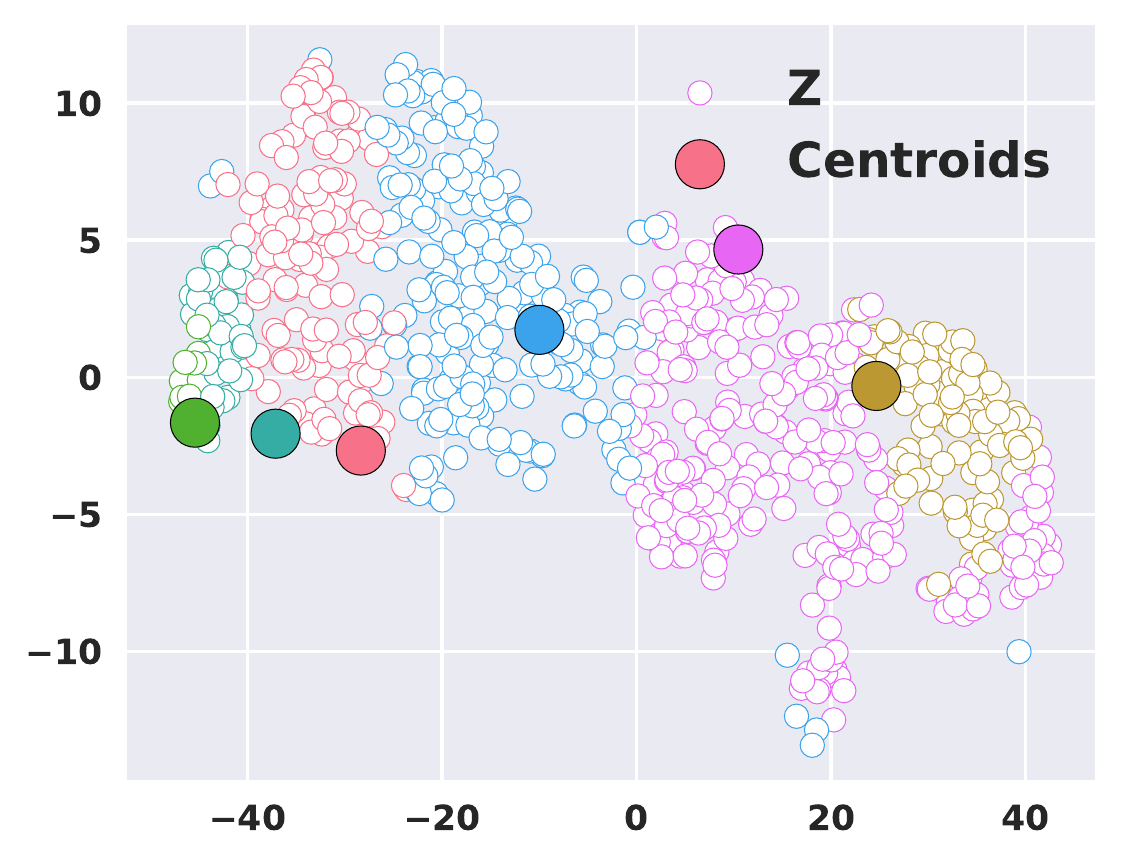}	&
			\includegraphics[width=0.32\linewidth, height=4cm]{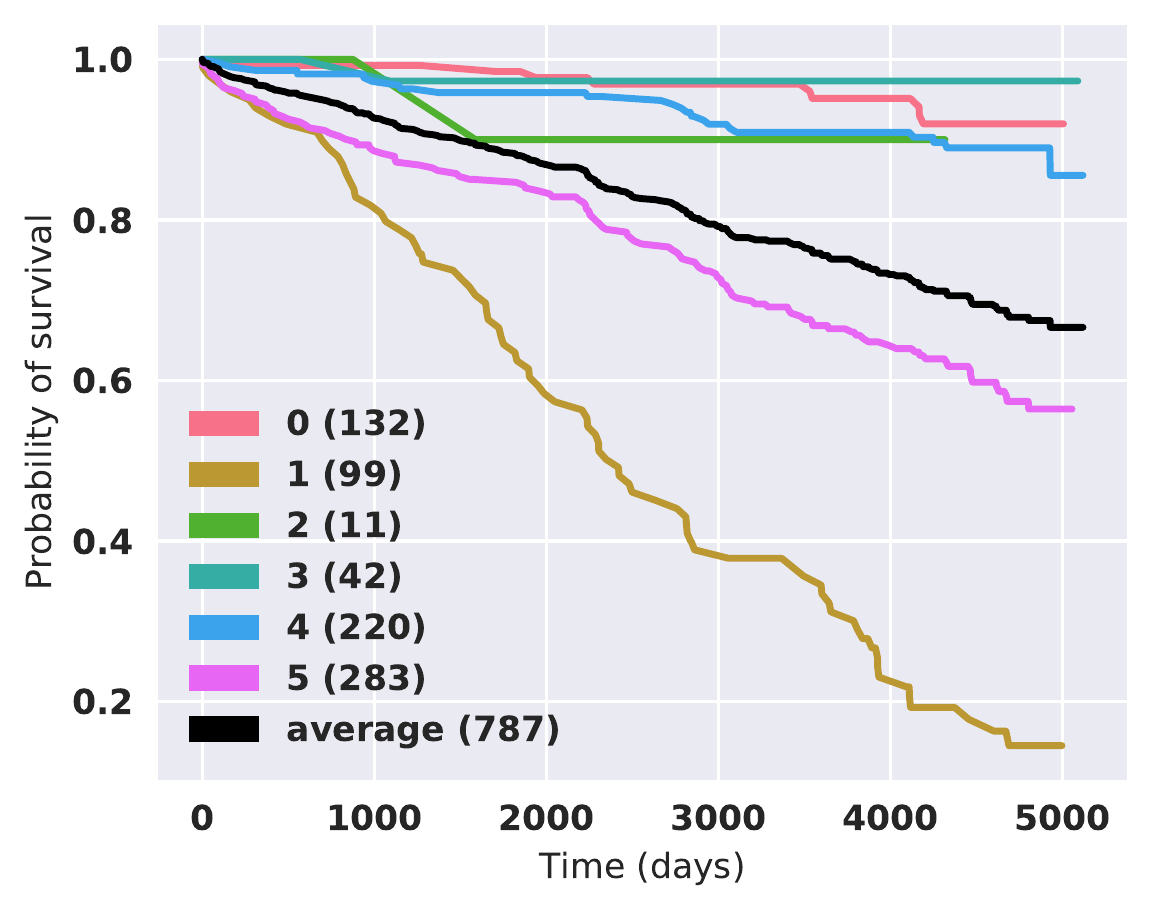}
			\\
			(a) Cluster assignments &
			(b) t-SNE plot  &
			(c)  Cluster specific Kaplan-Meir Curves 
		\end{tabular}		
		\caption{Inferred clusters on the testing set of {\sc flchain} dataset, with $K=25$ and $\gamma_{o}=4$ with corresponding individual probability distribution $q(\bs{\pi} | \CD)$.
		}
		\label{fg:flchain}
	\end{figure*}

\begin{figure*}[ht!]
		\centering
		\begin{tabular}{ccc}
			\includegraphics[width=0.32\linewidth,  height=5cm]{sca_figures/sleep/sleep_cluster_assignment.pdf}	&
			\includegraphics[width=0.32\linewidth,  height=4cm]{sca_figures/sleep/sleep_cluster_scatter.pdf}	&
			\includegraphics[width=0.32\linewidth, height=4cm]{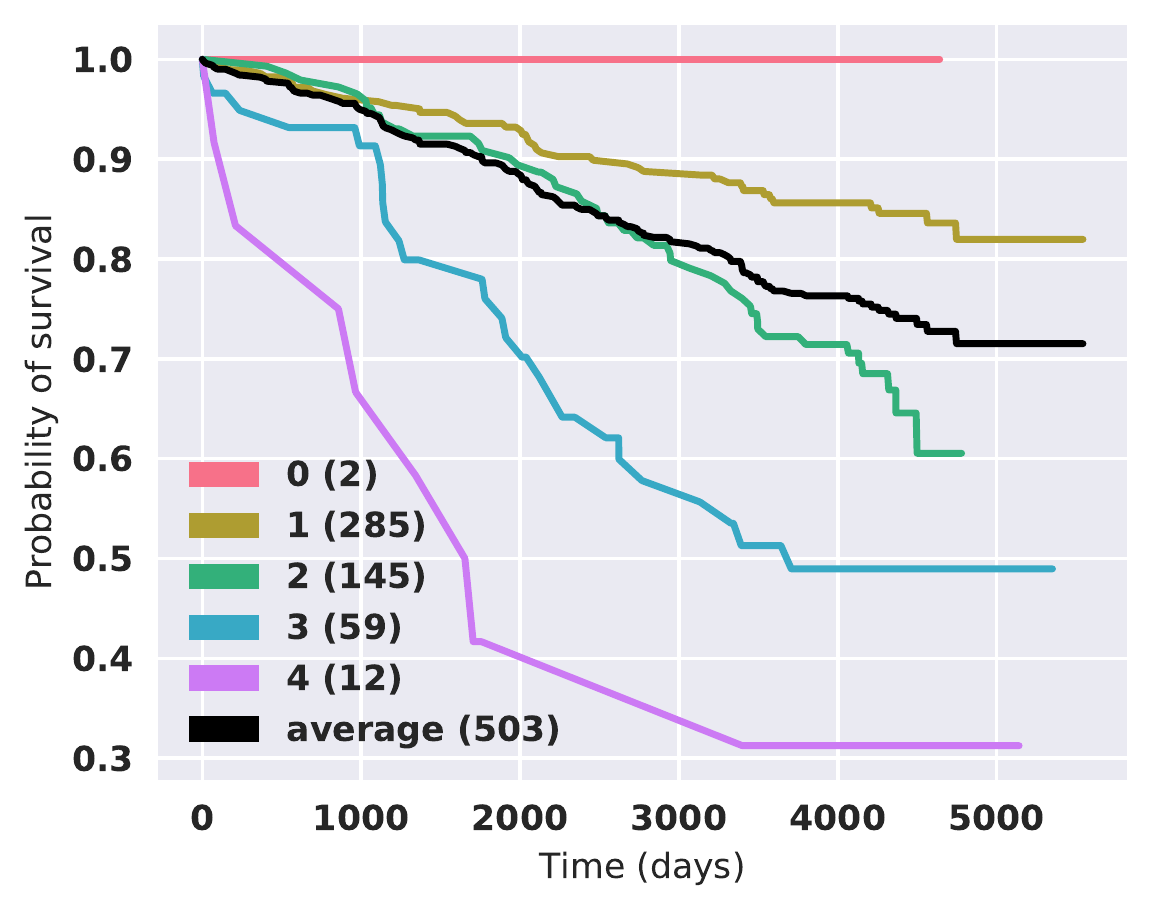}
			\\
			(a) Cluster assignments &
			(b) t-SNE plot  &
			(c)  Cluster specific Kaplan-Meir Curves 
		\end{tabular}		
		\caption{Inferred clusters on the testing set of {\sc sleep} dataset, with $K=25$ and $\gamma_{o}=3$ with corresponding individual probability distribution $q(\bs{\pi} | \CD)$.
		}
		\label{fg:sleep}
	\end{figure*}

\begin{figure*}[ht!]
		\centering
		\begin{tabular}{ccc}
			\includegraphics[width=0.32\linewidth,  height=4cm]{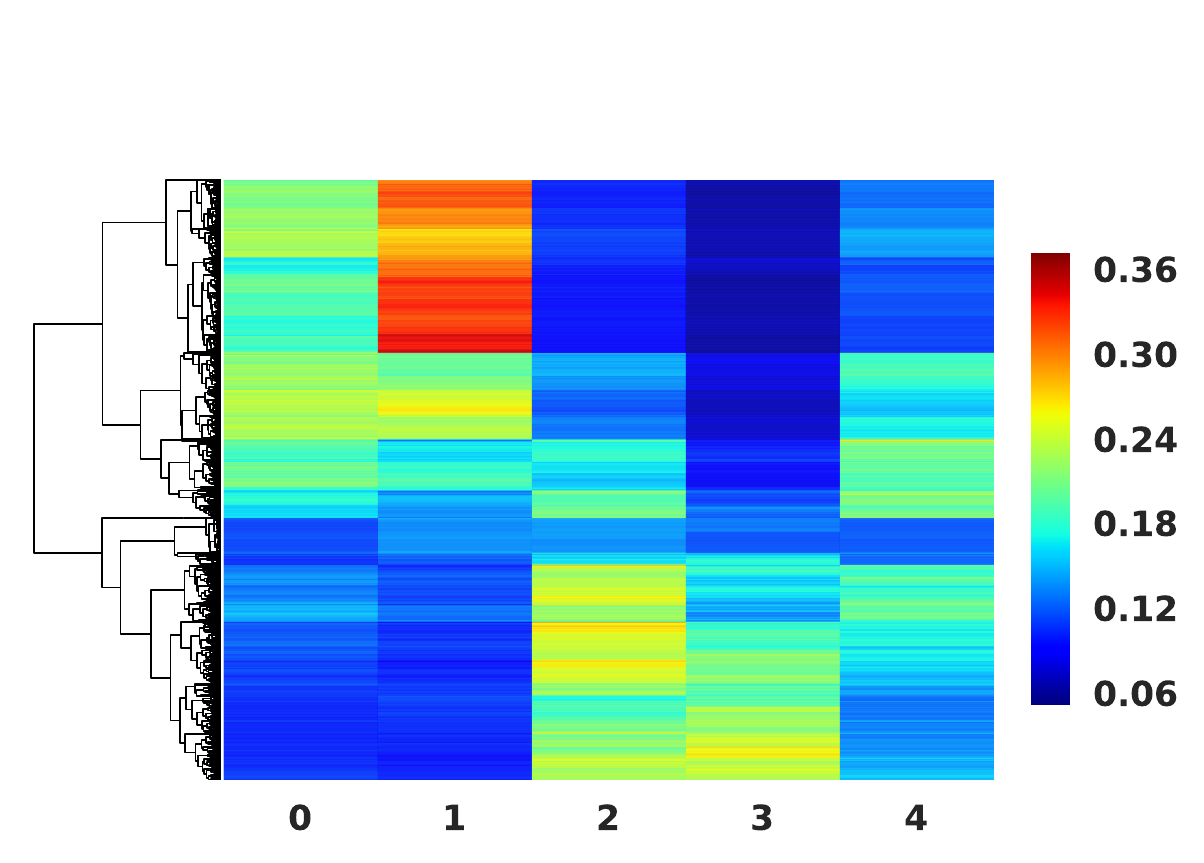}	&
			\includegraphics[width=0.32\linewidth,  height=4cm]{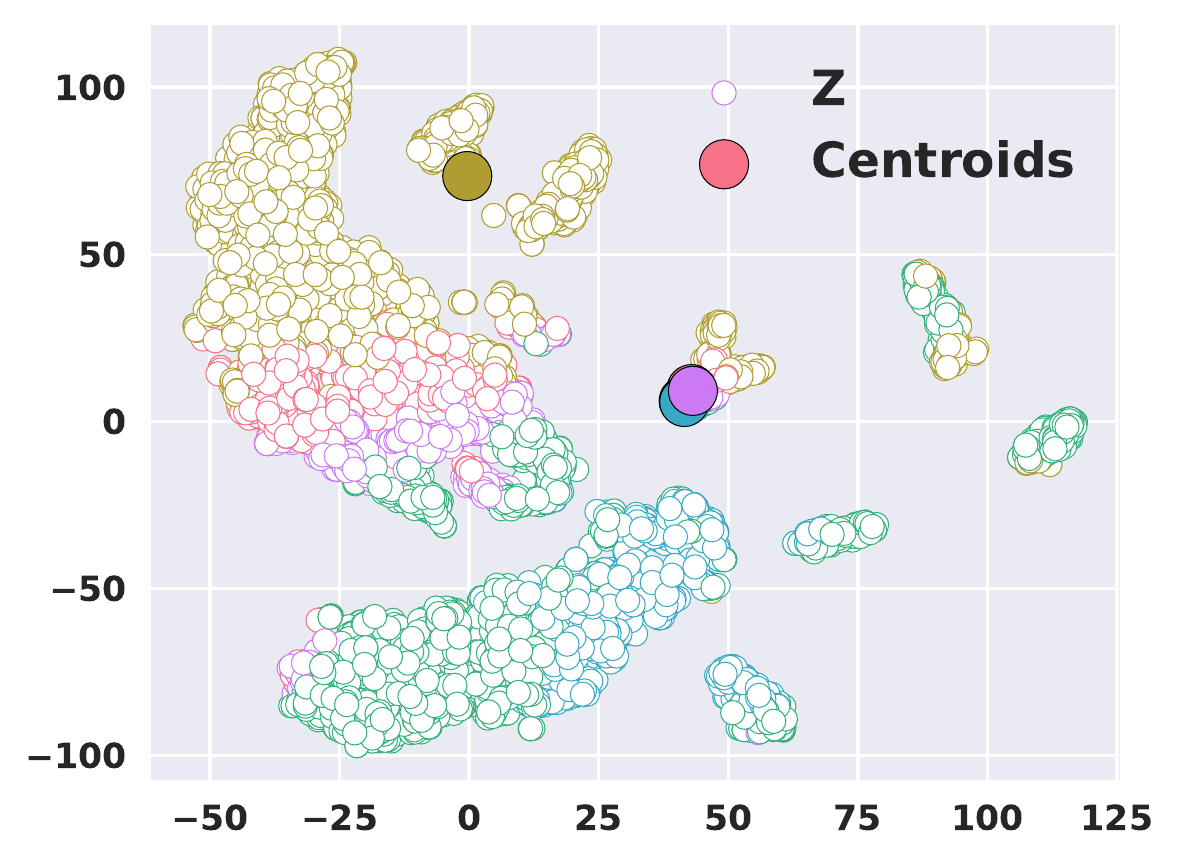}	&
			\includegraphics[width=0.32\linewidth,  height=4cm]{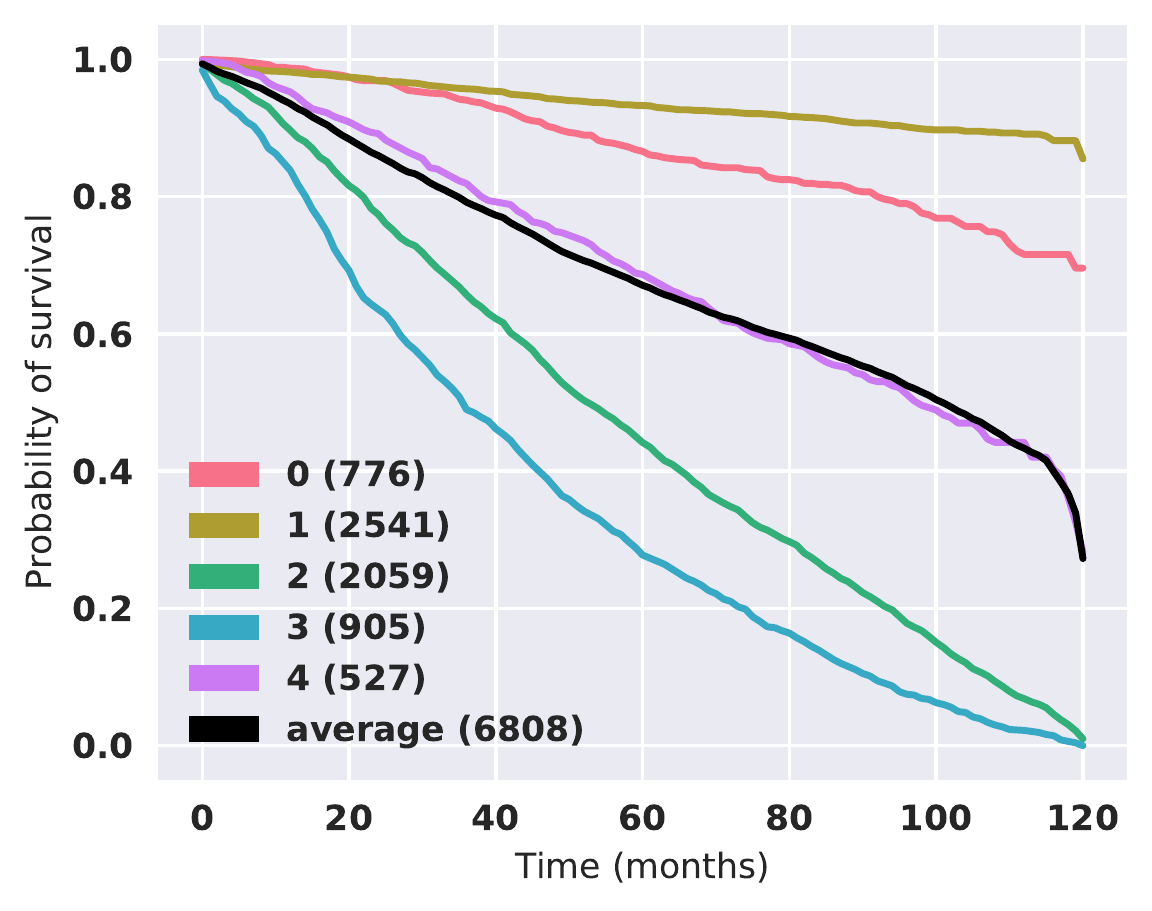}
			\\
			(a) Cluster assignments &
			(b) t-SNE plot  &
			(c)  Cluster specific Kaplan-Meir Curves 
		\end{tabular}		
		\caption{Inferred clusters on the testing set of {\sc seer} dataset, with $K=25$ and $\gamma_{o}=2$ with corresponding individual probability distribution $q(\bs{\pi} | \CD)$.
		}
		\label{fg:seer}
	\end{figure*}

\begin{figure*}[ht!]
		\centering
		\begin{tabular}{ccc}
			\includegraphics[width=0.32\linewidth,  height=4cm]{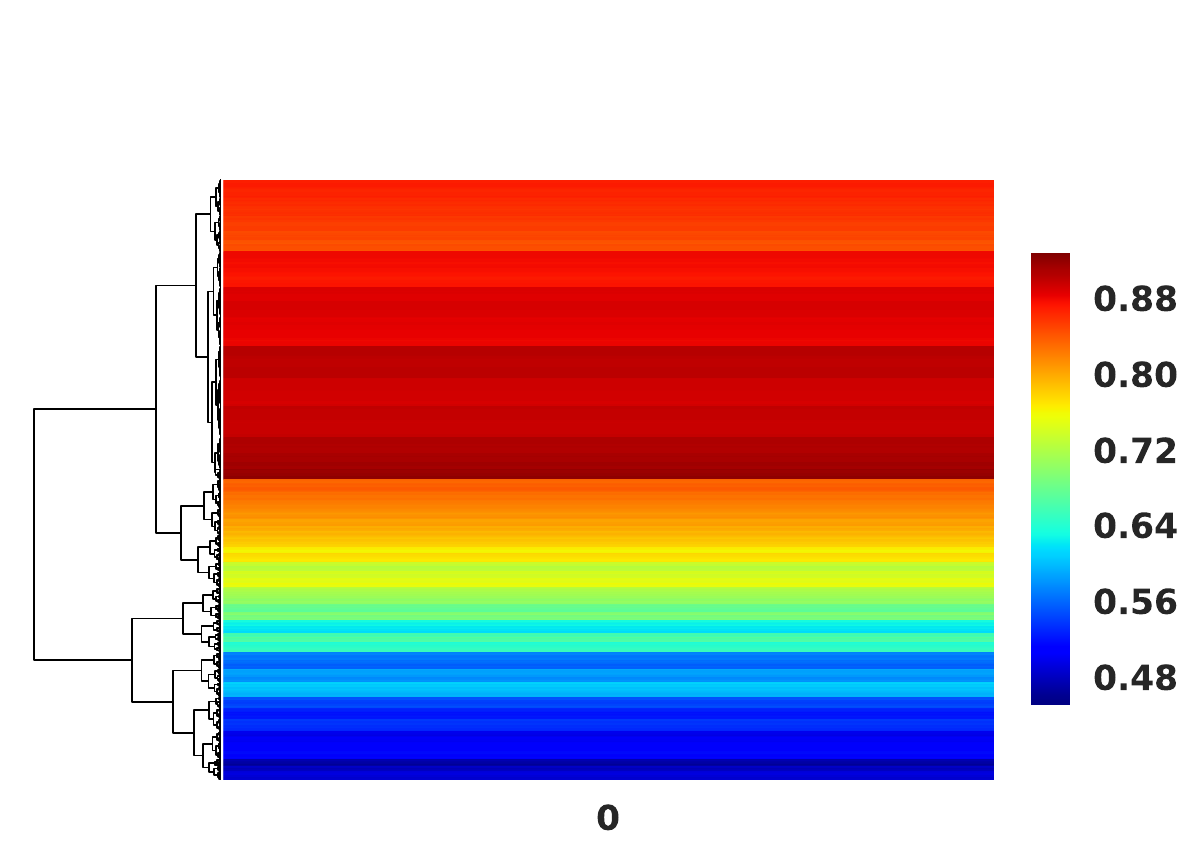}	&
			\includegraphics[width=0.32\linewidth,  height=4cm]{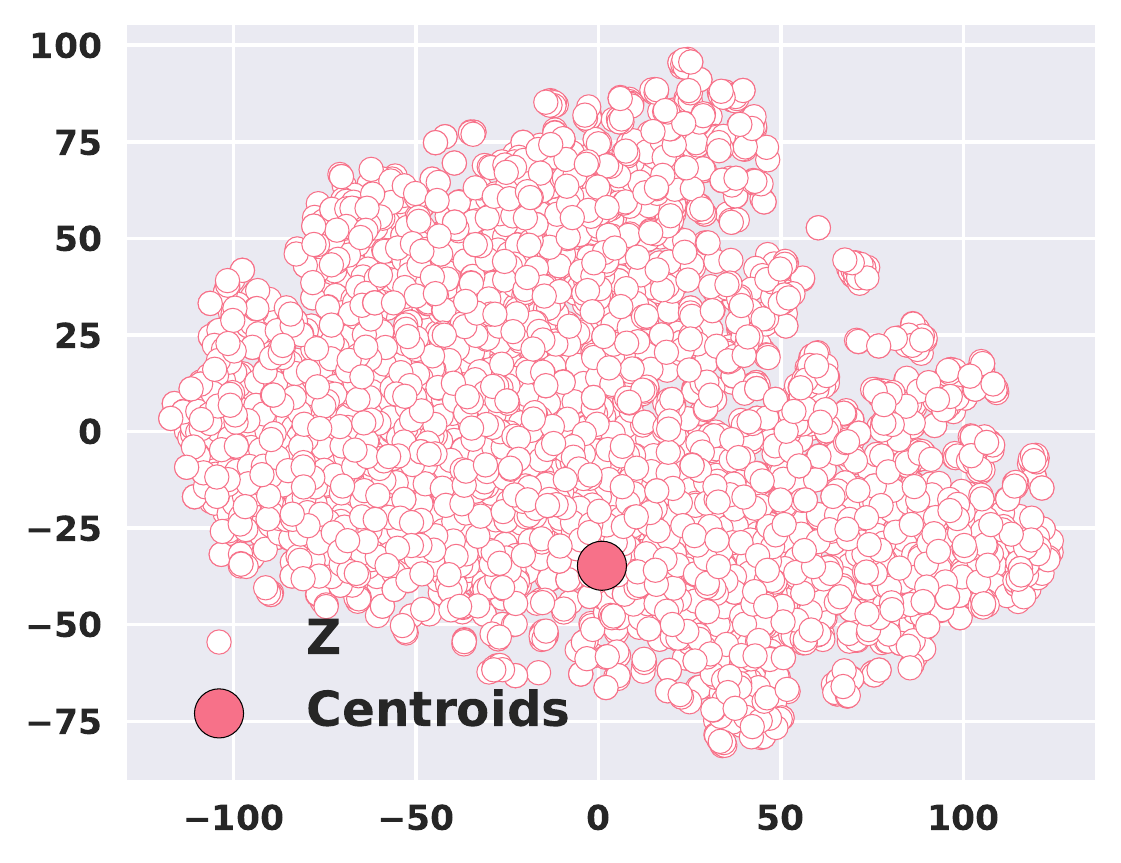}	&
			\includegraphics[width=0.32\linewidth, height=4cm]{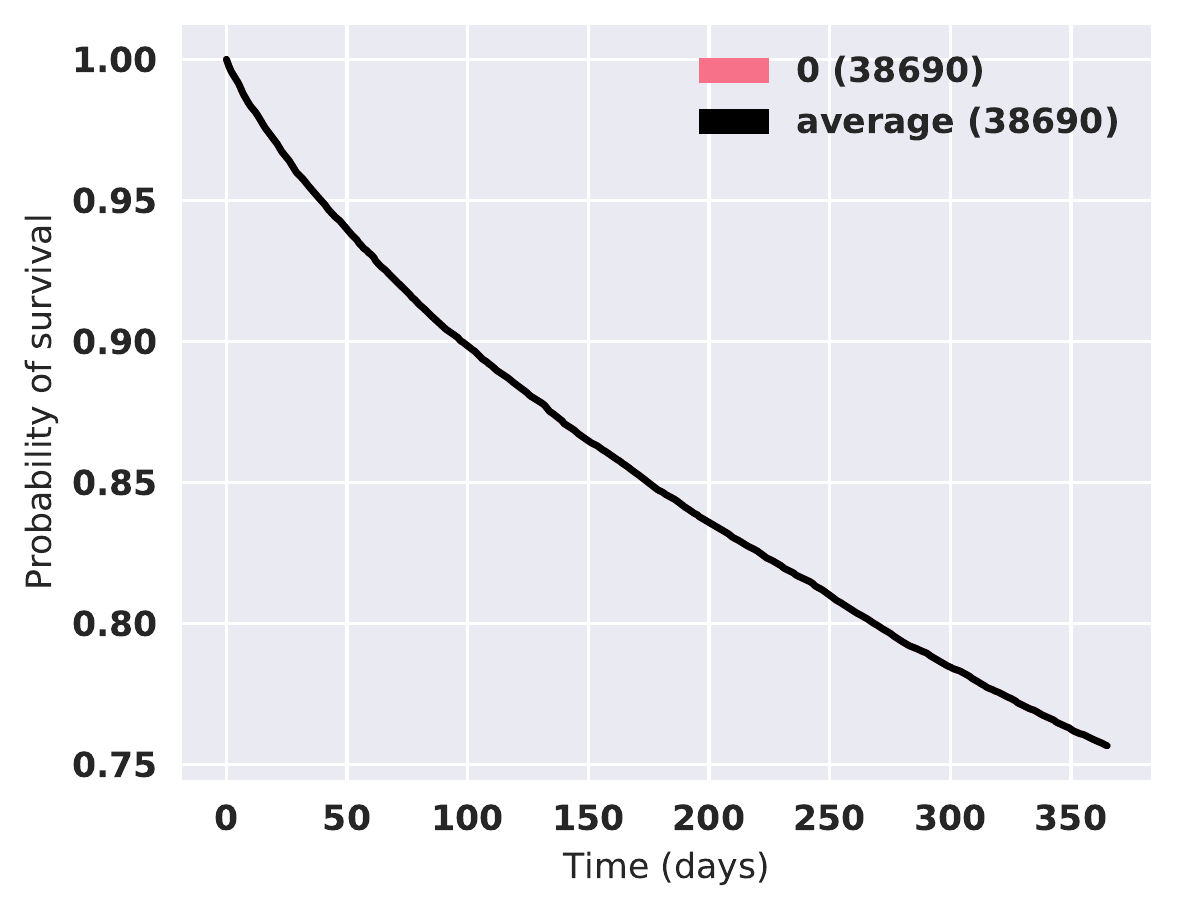}
			\\
			(a) Cluster assignments &
			(b) t-SNE plot  &
			(c)  Cluster specific Kaplan-Meir Curves 
		\end{tabular}		
		\caption{Inferred clusters on the testing set of {\sc ehr} dataset, with $K=25$ and $\gamma_{o}=2$ with corresponding individual probability distribution $q(\bs{\pi} | \CD)$.
		}
		\label{fg:sanofi}
	\end{figure*}

	We provide all the qualitative visualization of the latent-space representation, namely, $a)$ estimated individualized cluster assignment probability distributions; $b)$ $t$-SNE plots of both the centroids $\bs{c}_k$ with $\bs{z}$; $c)$ cluster-specific Kaplan Meir curves. Refer to Figures \ref{fg:pooled} - \ref{fg:sanofi} for SCA results on {\sc framingham}, {\sc support}, {\sc flchain}, {\sc sleep}, {\sc seer} and {\sc ehr} datasets respectively.

	\begin{table*}[ht!]
		\vspace{3mm}
		\centering
		\caption{SCA notations.}
		\label{tb:notation}
		
		\begin{tabular}{ll}
			Notation & Description \\
			\toprule
			$\xv_i$& vector of covariates  for subject $i$\\
			$t_i$  & empirical time-to-event (censored or non-censored)\\
			$l_i$ & censoring indicator, where $l_i =1$ represents a non-censored event \\
			$f(t| \xv)$ & time density function\\
			$F(t| \xv)$ & cumulative density function\\
			$S(t| \xv)$ & survival function, defined as $1-F(t| \xv)$\\
			$h(t| \xv)$ & hazard rate (risk score) function\\
			$r_{\bs{\psi}}(\xv)$ & deterministic encoder mapping, parametrized by $\bs{\psi}$\\
			$g_{\thetav}(\zv, \epsilonv)$ & stochastic mapping for synthesizing event times, parametrized by $\bs{\theta}$, where $\epsilonv$  is the source of stochasticity \\
			$st(\bs{c}_{u_n},\nu)$ & Student's $t$-distribution, with  $\nu$ degrees of freedom,  means $\{\bs{c}_k\}_{k=1}^\infty$ and $u_n$ mixture-component indicator for $\zv_n$ \\
			$\pi_k$ & mixture proportions\\
			$q(\bs{\pi}|\bs{\xi},\{\bs{x}_1\}_{n=1}^M) = {\rm Dir}(\bs{\pi}|\bs{\xi})$ & MLE based distribution for mixture assignments and proportions\\
			$	p(\bs{\pi}|\bs{\gamma},\{\bs{x}_n\}_{n=1}^M) = {\rm Dir}(\bs{\pi}|\bs{\gamma})$ & DP based distribution for mixture assignments and proportions\\
			\bottomrule
		\end{tabular}
	\end{table*}
	
	\begin{table*}[ht!]
		\vspace{3mm}
		\centering
		\caption{\small Performance metrics. SCA is the proposed model.}
		\label{tb:ci_rae_cov}
		\resizebox{\linewidth}{!}{
			\begin{tabular}{lrrrrrr}
				& {\sc ehr} & {\sc flchain} & {\sc support} & {\sc seer}  & {\sc sleep} &  {\sc framingham}\\
				\toprule
				RAE (non-censored, censored)  & & & & & &\\
				\hdashline
				DATE   & (\textbf{0.6107}, 0.2148) & (0.5222, 0.1159) & (0.6691, 0.1471)& (0.5289, 0.0294)& (0.5224, 0.2939) & (0.5122, 0.1888)\\
				DRAFT   &(0.7099, 0.1195) & (0.6399, 0.0445)& (0.7109,  \textbf{0.0418})& (0.6097, \textbf{0.0139})&  (0.7465, 0.3317)& (0.6697, 0.1119)\\
				S-CRPS  &(0.7240, 0.0773) & (0.6378, \textbf{0.0321})& (\textbf{0.4851}, 0.2333)& (0.5323, 0.0427)& (0.7330, \textbf{0.0519}) & (0.8369, \textbf{0.0108}) \\
				CoxPH   & -&- & -& -&-&- \\
				MTLR  & -&- & -& -&-&- \\
				SFM   &(0.6146, 0.2066) & (\textbf{0.5111}, 0.1313)& (0.6398, 0.3274)& (0.5294, 0.0329)& (\textbf{0.5162}, 0.2391)&  (\textbf{0.5074}, 0.2070)\\
				SCA  & (0.6186, 0.1897 )&(0.5134, 0.1199) &(0.6295, 0.2671) & (\textbf{0.5193}, 0.0480)& (0.5424, 0.2244)& (\textbf{0.5074}, 0.1819)\\				
				\toprule
				Mean Cov  & & & & & &\\
				\hdashline
				DATE  & \textbf{0.2477} & \textbf{0.3585} & 0.2987& \textbf{0.1485} & 0.5168 & \textbf{0.3624}\\
				DRAFT  &  5.0305 &   6.2952  & 3.8689 &  3.4501 & 8.4918   & 3.9911\\
				S-CRPS &0.8585&0.9412& 0.7351& 0.6036 & 1.0240 & 0.6225\\
				CoxPH &-&-&-&-&-&-\\
				MTLR&-&-&-&-&-&-\\
				SFM   & 0.2953 & 0.4484 &  0.3930 &  0.1993 & 0.5045 &0.3696\\
				SCA & 0.2842& 0.4458& \textbf{0.2891} & 0.2154& \textbf{0.4619} & 0.3870\\
				\toprule
				C-Index  & & & & & &\\
				\hdashline
				DATE  & 0.7756 & 0.8264&0.8421 & \textbf{0.8320}& 0.7416 &  0.7048\\
				DRAFT  &  0.7796 &0.8341 & 0.8560 & 0.8310 & \textbf{0.7617} &0.7005\\
				S-CRPS & 0.7704& 0.8286& \textbf{0.8685} & 0.8298 & 0.7529 & 0.7015\\
				CoxPH &0.7542& \textbf{0.8344}&0.8389& 0.8223&0.6435 & \textbf{0.7603}\\
				MTLR&-&-&-&-&-&-\\
				SFM   &  0.7786& 0.8318 &  0.8319 &  0.8314  &  0.7491  &0.7009\\
				SCA & \textbf{0.7809}& 0.8330& 0.8465 & 0.8306& 0.7498& 0.7072\\
				\bottomrule
				\label{tb:add_results}
			\end{tabular}
		}
	\end{table*}
	
	\FloatBarrier
	
	\bibliographystyle{ACM-Reference-Format}
	\bibliography{sca}